\documentclass[runningheads]{llncs}

\usepackage{eccv}

\usepackage[dvipsnames]{xcolor}

\newcommand{\bc}{\mathbf{c}}
\newcommand{\bd}{\mathbf{d}}

\newcommand{\bo}{\mathbf{o}}
\newcommand{\bp}{\mathbf{p}}

\newcommand{\br}{\mathbf{r}}
\newcommand{\bs}{\mathbf{s}}

\newcommand{\bx}{\mathbf{x}}

\newcommand{\bSigma}{\boldsymbol{\Sigma}}

\newcommand{\cG}{\mathcal{G}}

\newcommand{\cP}{\mathcal{P}}

\makeatletter
\DeclareRobustCommand\onedot{\futurelet\@let@token\@onedot}
\def\@onedot{\ifx\@let@token.\else.\null\fi\xspace}
\def\eg{e.g\onedot} 
\def\ie{i.e\onedot}

\def\etal{et~al\onedot}

\makeatother

\definecolor{yellow}{rgb}{1, 1, 0.7}
\definecolor{orange}{rgb}{1, 0.85, 0.7}
\definecolor{tablered}{rgb}{1, 0.7, 0.7}
\definecolor{red}{rgb}{1, 0, 0}

\definecolor{wincolor}{rgb}{0.85, 0.0, 0.0}

\definecolor{darkyellow}{rgb}{0.8, 0.8, 0.5}
\definecolor{darkred}{rgb}{0.7, 0.3, 0.3}
\definecolor{darkgreen}{rgb}{0.3, 0.7, 0.3}
\definecolor{blue}{rgb}{0.251, 0.498, 0.824}
\definecolor{green}{rgb}{0, 1.0, 0}
\definecolor{pink}{rgb}{1, 0.4, 0.7}

\definecolor{realred}{rgb}{0.95, 0.1, 0.0}

\usepackage{eccvabbrv}

\usepackage{graphicx}
\usepackage{booktabs}

\usepackage{colortbl}
\usepackage{multirow}

\usepackage{wrapfig}
\usepackage{makecell} 

\usepackage{algorithm,algpseudocode}

\usepackage[accsupp]{axessibility}  

\usepackage{hyperref}

\usepackage{orcidlink}

\definecolor{tabthird}{rgb}{1, 1, 0.7}
\definecolor{tabsecond}{rgb}{1, 0.85, 0.7}
\definecolor{tabfirst}{rgb}{1, 0.7, 0.7}

\newcommand{\wb}[1]{\textcolor[rgb]{0, 0, 0}{#1}}

\begin{document}

\title{Mini-Splatting: Representing Scenes with a Constrained Number of Gaussians} 

\titlerunning{Mini-Splatting}

\author{Guangchi Fang\orcidlink{0009-0001-2042-1182} \and Bing Wang\orcidlink{0000-0003-0977-0426}\thanks{Corresponding author}}

\authorrunning{G. Fang and B. Wang$^{\star}$}

\institute{Spatial Intelligence Group, The Hong Kong Polytechnic University
\email{guangchi.fang@gmail.com, bingwang@polyu.edu.hk}}

\maketitle

\begin{figure*}[h]
	\centering
	\includegraphics[width=1\linewidth]{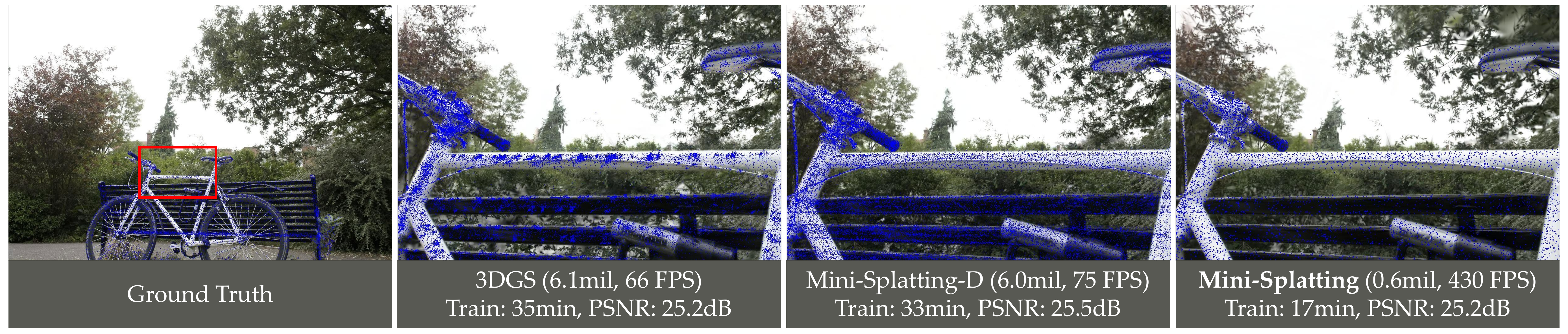}
    \caption{By reorganizing the spatial distribution of 3D Gaussians \cite{kerbl20233d}, our Mini-Splatting reduces the number of Gaussians (in millions) while enhancing model performance in terms of rendering speed, training time, and rendering quality. The Gaussian centers of a foreground object \textit{bicycle} are projected onto the rendered image as blue points, demonstrating the more uniform spatial distribution achieved by our algorithm.}
    
	\label{fig:teaser}
\end{figure*}

\vspace{-0.9cm}
\begin{abstract}


In this study, we explore the challenge of efficiently representing scenes with a constrained number of Gaussians. Our analysis shifts from traditional graphics and 2D computer vision to the perspective of point clouds, highlighting the inefficient spatial distribution of Gaussian representation as a key limitation in model performance. To address this, we introduce strategies for densification including blur split and depth reinitialization, and simplification through intersection preserving and sampling. These techniques reorganize the spatial positions of the Gaussians, resulting in significant improvements across various datasets and benchmarks in terms of rendering quality, resource consumption, and storage compression. Our Mini-Splatting integrates seamlessly with the original rasterization pipeline, providing a strong baseline for future research in Gaussian-Splatting-based works. 
\href{https://github.com/fatPeter/mini-splatting}{Code is available}.

  \keywords{Gaussian Splatting \and Point Clouds \and Scene Representation}
\end{abstract}


\section{Introduction}
\label{sec:intro}


Recent advancements in 3D Gaussian Splatting (3DGS) \cite{kerbl20233d} have showcased its significant potential across a range of applications, such as immersive rendering and 3D reconstruction, attributed to its capabilities for real-time and high-quality rendering. Unlike traditional radiance field-based algorithms \cite{mildenhall2020, barron2021mipnerf, barron2022mipnerf360}, 3DGS employs millions of elliptical Gaussians for scene modeling (\eg, \wb{1 to 6 million Gaussians for each scene} in the Mip-NeRF360 dataset \cite{barron2022mipnerf360}). The detailed characterization of these Gaussians, including their 3D position, opacity, covariance, and spherical harmonic (SH) coefficients, allows for a remarkably detailed and realistic scene representation. However, the extensive use of Gaussians also leads to inefficiencies, as depicted in Fig. \ref{fig:teaser}, where Gaussian centers are prone to clustering together. This suboptimal spatial distribution can constrain the rendering quality and speed. In this regard, our study aims to address the challenge of efficiently representing a scene with a constrained number of Gaussians.


The optimization of minimal Gaussian representation remains a relatively underexplored topic, with only a handful of compression-focused studies venturing into this area. Drawing inspiration from voxel pruning techniques used in grid-based neural representations \cite{yu2021plenoxels, li2023compressing}, Niedermayr \etal \cite{niedermayr2023compressed} proposed pruning Gaussians based on parameter sensitivity derived from image gradients. Similarly, Lee \etal \cite{lee2023compact} and LightGaussian \cite{fan2023lightgaussian} have implemented pruning strategies based on Gaussian opacity and scale. However, these methods, which primarily aim at storage compression, often neglect the inefficient spatial distribution of input Gaussians. As a result, direct pruning, as practiced in these approaches, tends to yield suboptimal simplification outcomes. Moreover, most existing 3D data simplification algorithms focus on preserving feature points \cite{katz2013improving, lang2020samplenet} or geometric structures \cite{pauly2002efficient, lv2022intrinsic}, making them less suitable for explicit representations with densely optimizable parameters like 3DGS. Therefore, adapting these techniques to the 3DGS model presents a non-trivial challenge.



In this paper, we start with a comprehensive analysis of Gaussian representation. By visualizing the projected Gaussian centers and rendered results, we find that 3D Gaussians are not uniformly distributed around the object surface, \wb{which limits the rendering speed and quality of the model}. To construct an efficient Gaussian representation, we pivot the task of directly simplifying 3D Gaussians to reorganizing their spatial positions employing our Gaussian densification and simplification method. Specifically, our proposed densification algorithm includes two key components: blur split, which involves splitting Gaussians in the blurring areas, and depth reinitialization, which uses merged depth points to reinitialize the scene. Our simplification approach comprises intersection preserving, which preserve Gaussians with maximal impact on the rendered image, and sampling, designed to preserve both the geometric structure and rendering quality of the original model. By stacking our densification and simplification techniques, we achieve a more uniform and efficient spatial distribution of Gaussians, as illustrated in Fig. \ref{fig:teaser}. Our main contributions are as follows:


\begin{itemize}
\setlength{\itemsep}{0pt}
\setlength{\parsep}{0pt}
\setlength{\parskip}{0pt}

    \item Through a deep analysis of the spatial distribution of Gaussians, we observe that the vanilla 3DGS presents two significant phenomena: `overlapping' and `under-reconstruction'. We identify that this inefficient distribution limits the rendering quality and speed, and it is non-trivial to achieve a minimal Gaussian representation while maintaining rendering quality.

    \item We propose a Gaussian densification and simplification algorithm to reorganize the spatial positions of Gaussians rather than directly pruning them. The proposed densification (blur split and depth reinitialization) and simplification (intersection preserving and Gaussian sampling) strategies, leveraging both screen-space and world-space information, effectively achieve a dense distribution of Gaussians around the object.


    \item By integrating densification and simplification with corresponding further processing, our proposed Mini-Splatting achieves a balanced trade-off between rendering quality, resource consumption, and storage. Extensive experimental results on multiple benchmarks and datasets demonstrate the potential and scalability of our method.


\end{itemize}

\section{Related Work}
\label{sec:related_work}

\subsection{Gaussian Splatting}

3D Gaussian Splatting (3DGS) \cite{kerbl20233d} represents a recent advancement in novel view synthesis, enabling real-time high-quality rendering. Over the past year, numerous follow-up works have emerged. The concept of 3DGS has been applied to domains such as autonomous driving \cite{yan2024street}, human avatar \cite{qian2023gaussianavatars}, content generation \cite{tang2023dreamgaussian, sun2024recent} and dynamic scene \cite{luiten2023dynamic}, resulting in advancements within each domain \cite{chen2024survey}. While several works aim to enhance 3DGS from the perspectives of graphics \cite{franke2024trips}, image processing \cite{yu2023mip}, and system performance \cite{durvasula2023distwar}, few studies focus on delving into the nature of Gaussians as the scene representation.

The Gaussian representation constitutes a key distinction between 3DGS and previous standard techniques for novel view rendering. While most prior works employ implicit \cite{mildenhall2020, barron2021mipnerf, barron2022mipnerf360} or explicit \cite{yu2021plenoxels, xu2022point, sun2022direct} neural representations with volume rendering based on ray marching, the primary challenge lies in the dense ray point sampling required by ray marching, which consequently limits both training and rendering speed. In contrast, 3DGS utilizes a large collection of elliptical Gaussians \cite{zwicker2002ewa} for scene representation and rasterize these Gaussians onto the image plane for alpha blending. This carefully designed rasterization pipeline circumvents the issues associated with ray marching. However, the performance of 3DGS models is constrained by the large number of Gaussians due to the rasterization process.

\subsection{3D Data Simplification}

For point cloud simplification purposes, conventional techniques such as farthest point sampling \cite{qi2017pointnet}, random sampling \cite{hu2020randla}, and grid sampling \cite{graham2017submanifold} are commonly utilized in point cloud processing tasks. In contrast, most works oriented towards point cloud simplification focus on preserving either feature points \cite{katz2013improving, chen2017fast, lang2020samplenet} or geometric structures \cite{pauly2002efficient, moenning2004intrinsic, lv2022intrinsic}. Several learning-based algorithms \cite{dovrat2019learning, lang2020samplenet, wen2023learnable} have been proposed to sample points during training, integrating downstream tasks in an end-to-end manner. However, it is non-trivial to directly apply these sampling techniques to 3DGS due to differences in representations. 

With recent advancements in neural radiance fields, several works have been proposed to simplify neural representations, primarily focusing on feature grids and introducing voxel pruning \cite{deng2023compressing, li2023compressing} or voxel masking \cite{xie2023hollownerf, rho2023masked} strategies. Inspired by these developments, similar pruning techniques have been adapted in recent 3DGS compression-oriented works \cite{niedermayr2023compressed, lee2023compact, fan2023lightgaussian}. For instance, Niedermayr et al. \cite{niedermayr2023compressed} proposed to prune Gaussians based on image gradients, while Lee et al. \cite{lee2023compact} and LightGaussian \cite{fan2023lightgaussian} performed pruning according to Gaussian opacity and scale. However, these works overlook the issue of inefficient spatial distribution of Gaussians, which can easily lead to suboptimal simplification results after Gaussian pruning. One concurrent work, RadSplat \cite{niemeyer2024radsplat}, also share a similar technical pipeline with our algorithm.

\section{Analyzing 3DGS from a Perspective of Point Clouds}
\label{sec:Background}
\newcommand{\RRR}{\mathbb{R}}

\subsection{Preliminaries of Gaussian Splatting}


\textbf{Gaussian Representation.} 3D Gaussian Splatting (3DGS) \cite{kerbl20233d} models the scene geometry as a set of 3D Gaussians $\{\cG_i | i=1, \cdots, N\}$. Each Gaussian $\cG_i$ is characterized by its opacity $\alpha_i\in[0,1]$, center $\bp_i \in \RRR^{3 \times 1}$ and covariance matrix in world space $\bSigma_i \in \RRR^{3\times3}$ \cite{zwicker2002ewa} as $\cG_i(\bx) = e^{-\frac{1}{2} (\bx-\bp_i)^T \bSigma_i^{-1}(\bx-\bp_i)}$. To render an image from these Gaussians, 3DGS sorts them in an approximate depth order based on the distance from their centers $\bp_i$ to the image plane, and further applies alpha-blending \cite{mildenhall2020} as follows:
\begin{equation}\label{eq:render} 
    \bc(\bx) = \sum_{i=1}^N w_i \cdot \bc_i, \quad \text{where } w_i=T_i\cdot \alpha_i \cdot \cG^{2D}_i(\bx), \quad T_i=\prod_{j=1}^{i-1} (1 - \alpha_j \cdot \cG^{2D}_j(\bx)),
\end{equation}
where $\bc_i$ represents the view-dependent color modeled by spherical harmonic (SH) coefficients, and $\cG^{2D}_i$ is the projected 2D Gaussian distribution of $\cG_i$ through a local affine transformation \cite{zwicker2002ewa}.

\noindent\textbf{Adaptive Density Control of Gaussians.} In 3DGS, sparse points from Structure-from-Motion (SfM) are first initialized as Gaussians. During the optimization, these Gaussians undergo an adaptive density control strategy, allowing the dynamic addition and removal of individual Gaussians. Specifically, Gaussians with a significant gradient are identified and are either split or cloned depending on their scales. Additionally, Gaussians with low opacity or an overlarge screen space size are pruned from the representation. Ablation studies presented in \cite{kerbl20233d} demonstrate that this clone and split strategy can control the density of Gaussians, which in turn benefits scene recovery.

\begin{figure*}[tb]
	\centering
	\includegraphics[width=1\linewidth]{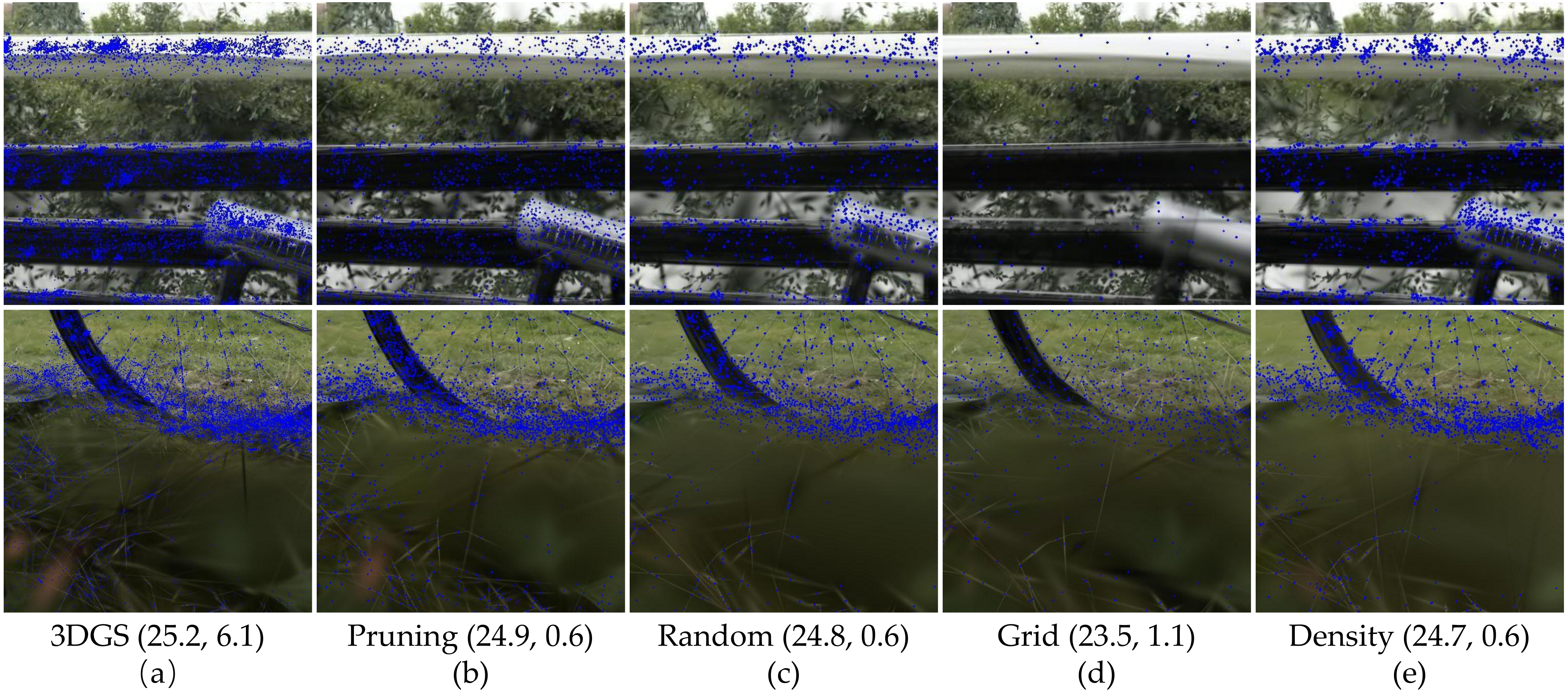}
    \vspace{-0.7cm}
    \caption{Visual analysis of Gaussian centers. (a): Visualization of the original 3DGS model's projected Gaussian centers, along with its corresponding rendering quality (PSNR in dB) and the number of Gaussians (in millions). Notably, we observe the phenomena of `overlapping' and `under-reconstruction'. (b), (c), (d), and (e): Visualization of Gaussian centers after applying different sampling techniques, including pruning, random sampling, grid sampling, and density-preserved sampling, respectively.}
	\label{fig:Visual_analysis}
\end{figure*}

\subsection{\wb{Challenges in Gaussian Simplification}}\label{sec:Challenges}

\noindent\textbf{Visual Analysis of Gaussians.} In 3DGS, scene geometry is represented using elliptical Gaussian primitives. Ideally, these Gaussians should be distributed around the geometric surface to accurately model the scene, particularly for the foreground Gaussians with relatively small 3D scales. Under this assumption, the Gaussian centers $\bp_i$, akin to point clouds, can be approximated as the geometric structure with their scales as the corresponding surface segments.

Following the experimental setup of 3DGS, we utilize sparse points as input initialization in the main paper. As depicted in Fig. \ref{fig:Visual_analysis} (a), we project Gaussian centers of the vanilla 3DGS model onto the rendered image as blue points. Typically, points captured by depth sensors or reconstructed by multiview stereo methods tend to uniformly adhere to the object surface. However, the first row illustrates the `overlapping' phenomenon, where most Gaussians are clustered in certain areas but sparse in neighboring parts. The second row highlights `under-reconstruction', wherein 3DGS fails to model the details of a specific area and presents artifacts in the form of Gaussian blurs. These two phenomena collectively indicate the inefficient spatial distribution of Gaussians, which underscores the inherent redundancy of the Gaussian representation. Additional analysis of 3DGS from initialization with dense point cloud is provided in the appendix.



\noindent\textbf{Gaussian Pruning and Sampling.} The most straightforward approach to reduce the number of Gaussians is to perform pruning and sampling. Similar to the importance calculation in \cite{li2023compressing}, we accumulate all blending weights $w_{ij}$ of Gaussian $\cG_i$ to obtain its importance score $I_i=\sum_{j=1}^K w_{ij}$, where $K$ is the total number of rays intersected with $\cG_i$. Then, we prune Gaussians with importance lower than a certain threshold. The Gaussian representation is further optimized until convergence and is then shown in Fig. \ref{fig:Visual_analysis} (b). The pruning strategy can reduce the number of Gaussians from 6.1 million to 0.6 million, but it also degrades PSNR from 25.2 dB to 24.9 dB, and the phenomena of `overlapping' and `under-reconstruction' still exist. We also explore combining pruning with various point cloud sampling techniques, such as random sampling, grid sampling, and density-preserved sampling, shown in Figs. \ref{fig:Visual_analysis} (c), (d), and (e), respectively. These results suggest that addressing the inefficient spatial distribution of Gaussians and achieving a minimal Gaussian representation while maintaining rendering quality requires further efforts.

\section{Methodology}
\label{sec:method}

\begin{figure*}[tb]
	\centering
	\includegraphics[width=1\linewidth]{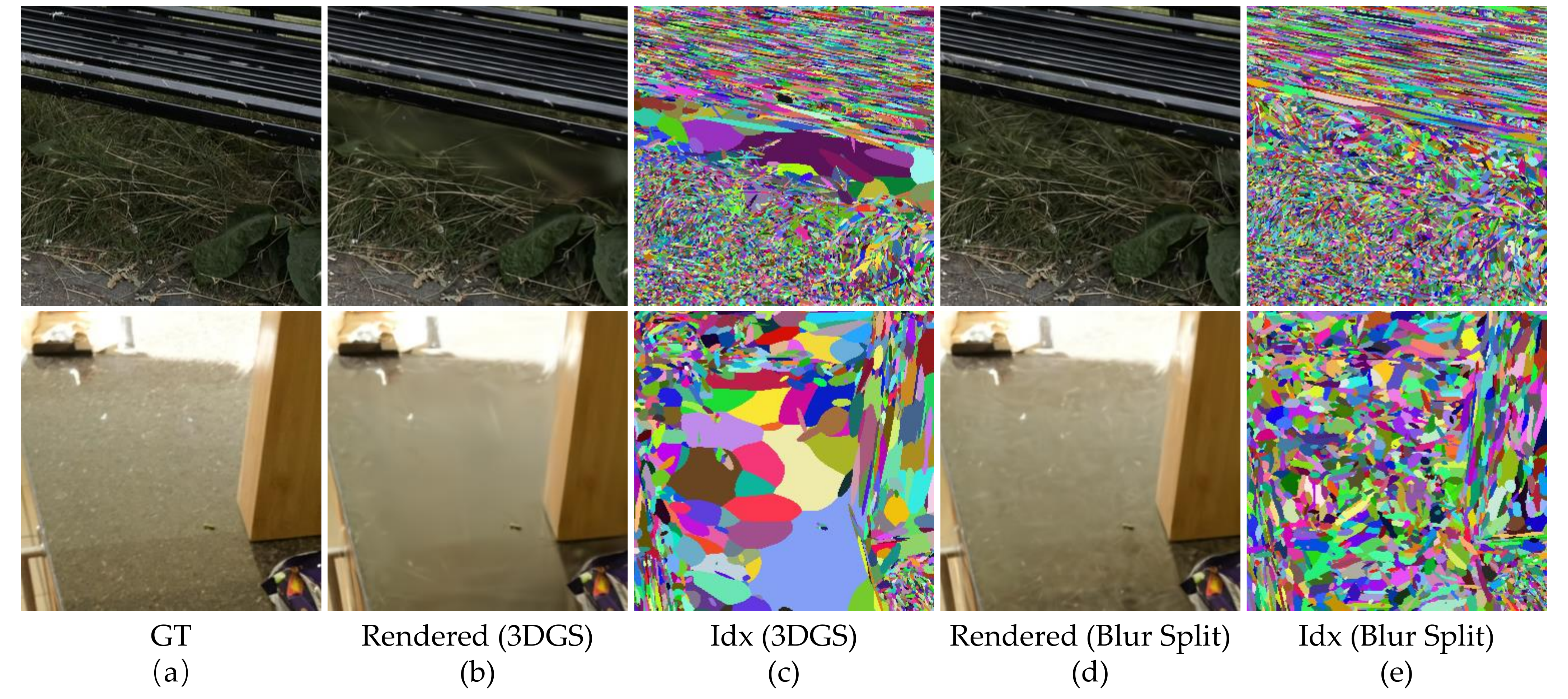}
    \vspace{-0.7cm}
    \caption{Visual analysis of blur split. (a) Ground truth image. (b) Rendered image using the original 3DGS model \cite{kerbl20233d}. (c) Rendered index of Gaussians with the maximum contribution to \wb{each pixel}. (d) Rendered image after applying the blur split. (e) Rendered Gaussian index after applying the blur split.} 
	\label{fig:blur_split}
\end{figure*}

In this section, we introduce our densification and simplification algorithms aimed at reorganizing the spatial distribution of Gaussians, thereby constraining the number of Gaussians while preserving rendering quality.

\subsection{Densification}\label{sec:Densification}

Our densification approach comprises two key components: blur split and depth initialization, to obtain Gaussians with a dense and uniform  spatial distribution.

\noindent\textbf{Blur Split.} The blur split strategy aims to address the Gaussian blur artifacts, as shown in Fig. \ref{fig:blur_split} (b), which are also discussed in Sec. \ref{sec:Challenges}. The gradient-based split and clone strategy \cite{kerbl20233d} may fail in areas with smooth color transitions, as illustrated in Fig. \ref{fig:blur_split} (a). Consequently, the corresponding oversized Gaussians $\cG_i$ tend to be preserved during optimization. One intuitive approach adopted by 3DGS is to prune Gaussians with an excessively large scale in screen space based on their projected 2D distribution $\cG^{2D}_i$. However, this pruning method does not address the issue of `under-reconstruction'.

As illustrated in Fig. \ref{fig:blur_split} (c), we render indexes of Gaussians with the maximum contribution to a pixel as $i_{max}=\arg\max\limits_{i} w_i$ (\ie, the Gaussian with the maximum weight in alpha-blending) onto the image. The rendered indexes $i_{max}(\bx)$ of each pixel $\bx$ are mapped to random colors. To distinguish with the rendered indexes from all Gaussians $i_{max}(\bx)$, we denote the projected index of $i$th Gaussian $\cG_i$ as $i(\bx)$. A key observation is that, compared to the projected index $i(\bx)$ (\ie, the projected area of Gaussian $\cG_i$), the blurry artifacts are more directly related to the rendered index $i_{max}(\bx)$. This indicates that Gaussians with large maximum contribution areas lead to the problem of `under-reconstruction'.

Based on this observation, for each image, we identify Gaussians with large blurry areas, denoted as $\cG^{blur}$, using the following criterion:
\begin{equation}\label{eq:blur_split} 
    \cG^{blur}=\{\cG_i | S_i > \mathcal{T}_{blur} \land i \in [1, N]\}, \quad \mathcal{T}_{blur}=\theta_{blur} \cdot H \cdot W.
\end{equation}
\wb{Here, $S_i=\sum_{\bx=(1,1)}^{(H, W)} \delta(i(\bx)=i_{max}(\bx))$ represents the maximum contribution area of Gaussian $\cG_i$, where $\delta$ is the indicator function, and $(H, W)$ denotes the image resolution. The value of $S_i$ can be computed during the forward pass of rasterization. The hyperparameter $\theta_{blur}$ is empirically set as $2 \times 10^{-4}$ to control the extent of the blurry area. Gaussians that meet the criteria $\cG^{blur}$ are then split according to the split strategy in \cite{kerbl20233d} during the optimization process. The results of this deblurring strategy are illustrated in Fig. \ref{fig:blur_split} (d) and (e), showcasing the effectiveness of the blur split approach in enhancing image clarity.}

\begin{figure*}[tb]
	\centering
	\includegraphics[width=1\linewidth]{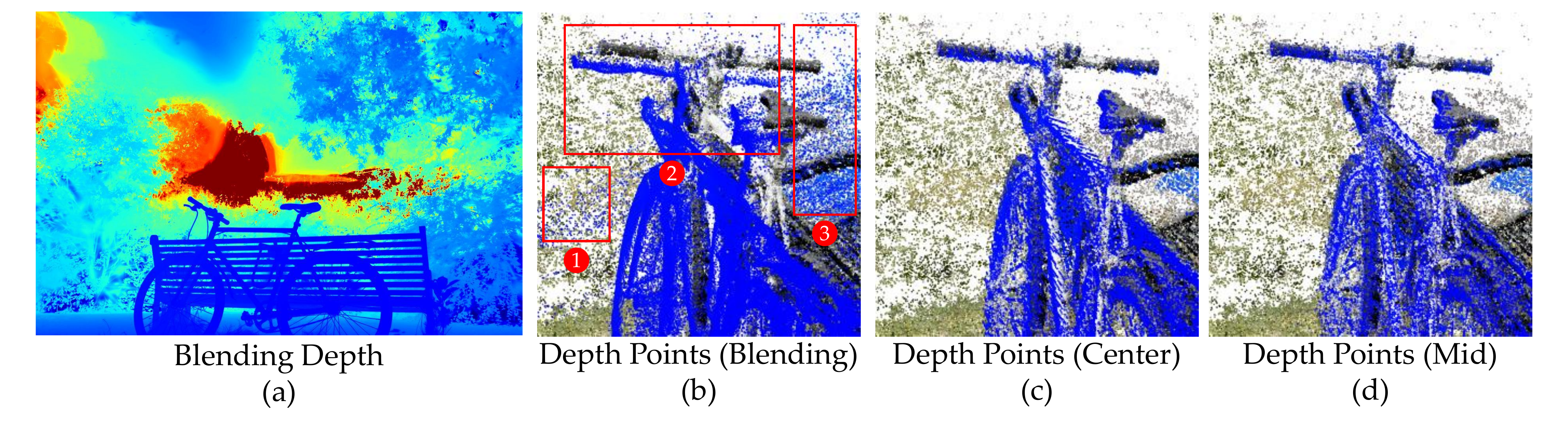}
    \vspace{-0.7cm}
    \caption{Visual analysis of depth map and depth points from 3DGS \cite{kerbl20233d}. (a): Rendered depth map from alpha blending. (b): Reconstructed depth points from alpha blending with red boxes highlighting significant artifacts. (c): Reconstructed depth points from our depth map using Gaussian centers. (d): Reconstructed depth points from our depth map using mid-points.}
	\label{fig:depth_points}
\end{figure*}

\noindent\textbf{Depth Reinitialization.} Our depth reinitialization leverages dense depth information to alleviate both the `overlapping' and `under-reconstruction' phenomena. The concept of utilizing depth to enhance neural rendering is both straightforward and effective, as evidenced by its application in several recent NeRF-based algorithms \cite{deng2022depth, xu2022point}. However, in Gaussian-Splatting-based works, the primary challenge lies not in how to utilize the reconstructed depth, but in how to generate accurate and practical depth data.


Similar to NeRF \cite{mildenhall2020}, 3DGS can render blending depth by directly replacing the Gaussian color $\bc_i$ with the depth of its center $d_i$, as $d^{blend} = \sum_{i=1}^N w_i \cdot d_i$. This approach is popular among many subsequent works, and the rendered depth map is shown in Fig. \ref{fig:depth_points} (a). However, we observe that this seemingly decent 2D depth map can be misleading, often resulting in a depth point cloud with significant artifacts. In Fig. \ref{fig:depth_points} (b), we reproject the blending depth to world space and also visualize Gaussian centers with their colors. Three types of artifacts (depth collapse, object misalignment, and blending boundary) are highlighted with red boxes for better illustration. These failures can be attributed to the alpha blending of multiple Gaussians, and more analysis is provided in the appendix.

\begin{figure*}[tb]
	\centering
	\includegraphics[width=1\linewidth]{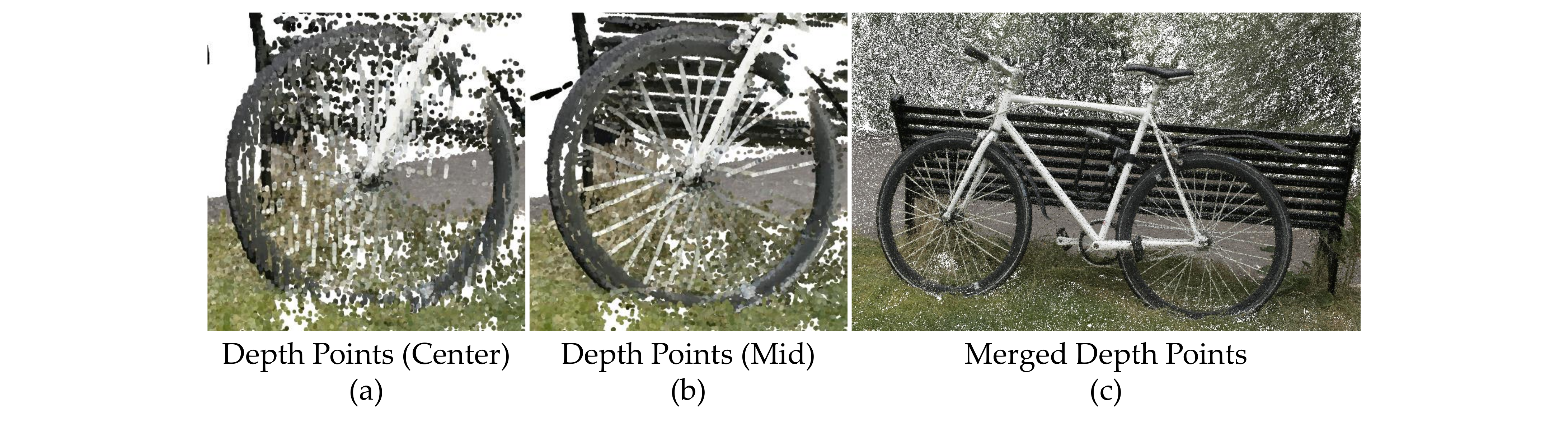}
    \vspace{-0.7cm}
    \caption{Depth points. (a): Depth points from our depth map using Gaussian centers. (b): Depth points from our depth map using mid-points. (c): Merged depth points from our depth map using mid-points.}
	\label{fig:merged_depth}
\end{figure*}

Here, we present our formulation of the rendered depth. Given one elliptical Gaussian defined by its scale $\bs=(s_x, s_y, s_z)$, we model the Gaussian as an ellipsoid in 3D space as $g(x, y, z)=\frac{x^2}{s_x^2} + \frac{y^2}{s_y^2} + \frac{z^2}{s_z^2} = 1$. With an input ray $\br(t)=\bo+t\bd$, where $\bo$ and $\bd$ are the ray origin and the ray direction transferred in the coordinate system of the ellipsoid, we can simply calculate the position of the mid-point between two intersection points of the ray and the ellipsoid. We set the depth of this mid-point as our Gaussian depth $d^{mid}_i$. Another intuitive formulation is to calculate the optimized $t$ where the maximum density of the Gaussian distribution along the ray is reached. We adopt the former formulation because these two share the same numerical result, but the former one can provide an auxiliary discriminant (\ie, the discriminant to distinguish the intersection point of the ray and the ellipsoid exists or not). Details of the calculation are provided in the appendix.

For multiple Gaussians, we only collect the Gaussian with the maximum contribution to the pixel to avoid the artifacts from alpha blending. The final depth $d^{mid}$ can be formulated as $d^{mid}=d^{mid}_{i_{max}}$, where $i_{max}=\arg\max\limits_{i} w_i$. In practice, we observe that setting the depth based on the Gaussian center, as $d^{center}=d_{i_{max}}$, can also achieve a similar rendering quality as $d^{mid}$ after further optimization. However, considering that $d^{mid}$ shows a better reconstruction result as illustrated in Figs. \ref{fig:merged_depth} (a) and (b), we chose to include it in our main pipeline of Mini-Splatting for potential future usage.

After generating the depth map, we reproject all depth points into the world space. For each image, we randomly select a certain number of points as initial points and assign the ground truth colors as their corresponding colors. The total number of sampled points is empirically set at around 3.5 million for each scene. During optimization, we repeatedly reinitialize our Gaussian representation with these points to reorganize the spatial distribution of Gaussians. The resulting merged point cloud is shown in Fig. \ref{fig:merged_depth} (c). These dense and uniform depth points assist our model in initializing a dense and uniform point distribution.



\subsection{Simplification}


After the densification process, we achieve a denser Gaussian representation of the scene. To further refine this representation, we propose simplification techniques, including intersection preserving and Gaussian sampling, to attain an efficient Gaussian representation while still maintaining high-quality rendering.

\noindent\textbf{Intersection Preserving.} Drawing inspiration from the concept of ray-mesh intersection, our intersection preserving technique is designed to discard Gaussians that do not directly intersect with the ray. However, we avoid strictly converting the smooth opacity or blending weights into binary values of ${0, 1}$, as this could potentially compromise the rendering quality.

In the depth reinitialization step, we model 3D Gaussians as ellipsoids and determine their depths based on the mid-points. We render the depth of the Gaussian with the maximum contribution, representing it as the depth of the scene. These mid-points, indicating the intersection of the rays with the scene representation, serve as the basis for our intersected Gaussians $\cG^{int}$. Specifically, for each image, we only retain Gaussians that provide these intersection points, and formulate these Gaussians as the intersected Gaussians:
\begin{equation}
    \cG^{int}=\{\cG_{i} | i \in \mathcal{I}_{max} \land i \in [1, N]\},
\end{equation}
where $\mathcal{I}_{max}$ is the collection of all indexes in the rendered indexes $i_{max}(\bx)$.


\setlength{\columnsep}{10pt}
\begin{wrapfigure}[12]{R}{0.46\textwidth}
\vspace{-0.3cm}
\centering
\raisebox{5pt}[\dimexpr\height-1\baselineskip\relax]{
\includegraphics[width=1\linewidth]{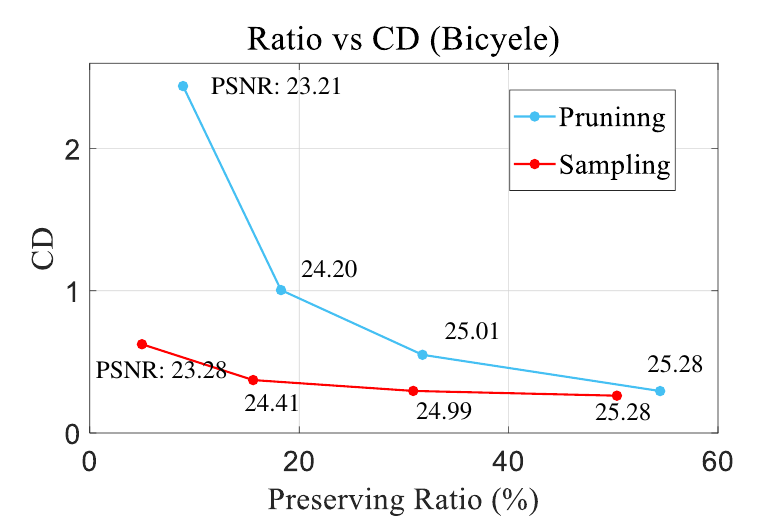}}
\vspace{-0.9cm}
\caption{The relation between the preserving ratio and the geometry.}
\label{fig:sampling}
\end{wrapfigure} 
\noindent\textbf{Importance-weighted Sampling.} In this section, we address the limitations of the direct pruning strategy and introduce our importance-weighted sampling approach. This method considers the geometric structure of the scene, thereby enhancing rendering quality while maintaining a sparse set of Gaussians.



The concept of pruning in neural representation is initially introduced for grid-based neural radiance fields \cite{yu2021plenoxels, li2023compressing}, aimed at removing empty or low-information voxels, and this technique can also be extended to Gaussian representations. By assigning each Gaussian a pre-defined importance value $I_i$, we can simply prune Gaussians with low importance. While this strategy is effective for removing a limited number of Gaussians, it may lead to a collapse in rendering quality when the pruning ratio is relatively large. The main issue lies in the fact that Gaussian importance only reflects its significance before simplification. Direct pruning can easily disrupt local geometry, which in turn degrades the final rendering quality after further optimization. Specifically, neighboring Gaussians often exhibit similar importance in a given area, causing them to be either removed or preserved simultaneously, thus risking the destruction of local geometry. To illustrate this point, we conduct Gaussian pruning on our intersected Gaussians and measure the geometry quality using the chamfer distance between the centers of pruned and non-simplified Gaussians. The significant chamfer distance observed after pruning, especially at low preserving ratios, suggests that relying solely on Gaussian importance leads to suboptimal simplification results.

Compared to deterministic pruning, a stochastic sampling strategy can better maintain overall geometry. Working under this observation, we incorporate Gaussian importance $I_i$ into the sampling probability of each Gaussian, formulated as $\cP_i=\frac{I_i}{\sum_{i=1}^N {I_i}}$. Thus, Gaussians can be sampled based on their sampling probability. As illustrated in Fig. \ref{fig:sampling}, the results demonstrate that this importance-weighted sampling strategy exhibits a stronger capability to preserve geometry quality, thereby enhancing the final rendering result. In our experiments, we find that using blending weight-based importance yields better performance in both pruning and sampling. Additionally, different combinations of blending weight and other information (\eg, the scale of projected Gaussians and the rendered index) exhibit varying performance on indoor and outdoor scans. Further implementation details of the importance metric are provided in the appendix.






\section{Implementation Variants}
\label{sec:applications}


Based on our densification and simplification algorithms, we construct three implementation variants. Details of each variant are provided in the appendix.


\noindent\textbf{Mini-Splatting.} Our Mini-Splatting is specifically designed for resource-efficient training and rendering. Following a similar optimization schedule as 3DGS \cite{kerbl20233d}, we set the total number of optimization steps to $30K$. Densification is enabled until the $15K$th iteration, after which simplification is conducted at the $15K$th and $20K$th iterations. Additionally, we observe that incorporating view-dependent colors barely enhance densification. Therefore, we only enable SH coefficients and increase the SH level after simplification ($15K$). 




\noindent\textbf{Mini-Splatting-D.} Our Mini-Splatting-D is tailored for quality-prioritized rendering. While its overall pipeline is similar to that of Mini-Splatting, the main distinction lies in the omission of the simplification process. 

\noindent\textbf{Mini-Splatting-C.} Our Mini-Splatting-C is designed for scenarios requiring storage compression. This pipeline is directly based on our Mini-Splatting. Given a pre-trained Mini-Splatting model, we store the Gaussian centers as \texttt{float32}, while other attributes, such as scales and SH coefficients, are processed using transform coding \cite{de2016compression} and a standard zip compression.







\section{Experiments}
\label{Sec:Experiments}


\noindent\textbf{Datasets.} We perform experiments on three real-world datasets: Mip-NeRF360 \cite{barron2022mipnerf360}, Tanks\&Temples \cite{Knapitsch2017} and Deep Blending \cite{hedman2018deep}. To ensure consistency and fairness in our evaluations, we adopt identical processing details for these datasets, including scene selection, train/test split, and image resolution, as specified in the official implementation of 3DGS \cite{kerbl20233d}.

\noindent\textbf{Implementation Details.} We implement our densification and simplification algorithms using PyTorch and integrate them into the optimization pipeline of 3DGS \cite{kerbl20233d}. We modify the Gaussian rasterization module of 3DGS to render Gaussian indexes and depth points. Further implementation details of our Mini-Splatting, Mini-Splatting-D, and Mini-Splatting-C are provided in the appendix.

\begin{table*}[tb]
    \renewcommand{\tabcolsep}{1pt}
    \centering
    \caption{Quantitative evaluation of our Mini-Splatting and previous works. 3DGS* indicates the retrained model from the official implementation.}
    \label{tab:quant}    
    \resizebox{1\linewidth}{!}{
    \begin{tabular}{@{}l@{\,\,}|cccc|cccc|cccc}
    Dataset & \multicolumn{4}{c|}{Mip-NeRF 360} & \multicolumn{4}{c|}{Tanks\&Temples} & \multicolumn{4}{c}{Deep Blending}  \\
    
  Method $|$ Metric & SSIM $\uparrow$ & PSNR $\uparrow$ & LPIPS $\downarrow$ & Num   & SSIM $\uparrow$ & PSNR $\uparrow$ & LPIPS $\downarrow$ & Num   & SSIM $\uparrow$ & PSNR $\uparrow$ & LPIPS $\downarrow$ & Num \\\hline

Plenoxels~\cite{yu2021plenoxels}         &    
                  0.626 &
 23.08 &                      0.463 &  -  &   
                   0.719 &
  21.08 &                      0.379 &  -  &  
                    0.795 &
   23.06 &                      0.510 &  -  \\INGP-Big~\cite{muller2022instant}        &    
                  0.699 &
 25.59 &                      0.331 &  -  &   
                   0.745 &
  21.92 &                      0.305 &  -  &  
                    0.817 &
   24.96 &                      0.390 &  -  \\mip-NeRF 360~\cite{barron2022mipnerf360} &    
                  0.792 & \cellcolor{tabsecond}27.69 &                      0.237 &  -  &   
                   0.759 &
  22.22 &                      0.257 &  -  &  
                    0.901 &
   29.40 &                      0.245 &  -  \\Zip-NeRF~\cite{barron2023zip}            & \cellcolor{tabsecond}0.828 &  \cellcolor{tabfirst}28.54 & \cellcolor{tabsecond}0.189 &  -  &   
                    -  &
 -  &                       -  &  -  &        
               -  &                       -  &                       -  &  -  \\ \hline
3DGS~\cite{kerbl20233d}                  &    
                  0.815 &
 27.21 &  \cellcolor{tabthird}0.214 & 3.36 &  
\cellcolor{tabthird}0.841 &
   23.14 &  \cellcolor{tabthird}0.183 & 1.78 &                      0.903 &
     29.41 & \cellcolor{tabsecond}0.243 & 2.98 \\
3DGS~\cite{kerbl20233d}*                 &    
                  0.815 &
 27.47 &                      0.216 & 3.35 & \cellcolor{tabsecond}0.848 &  \cellcolor{tabfirst}23.66 & \cellcolor{tabsecond}0.176 & 1.84 &  \cellcolor{tabthird}0.904 &  \cellcolor{tabthird}29.54 &  \cellcolor{tabthird}0.244 & 2.82 \\ \hline
Mini-Splatting-D                         &  \cellcolor{tabfirst}0.831 &  \cellcolor{tabthird}27.51 &  \cellcolor{tabfirst}0.176 & 4.69 &  
\cellcolor{tabfirst}0.853 & \cellcolor{tabsecond}23.23 &  \cellcolor{tabfirst}0.140 & 4.28 & \cellcolor{tabsecond}0.906 & \cellcolor{tabsecond}29.88 &  \cellcolor{tabfirst}0.211 & 4.63 \\
Mini-Splatting                           &  \cellcolor{tabthird}0.822 &
 27.34 &                      0.217 & 0.49 &  
                    0.835 &  \cellcolor{tabthird}23.18 &                      0.202 & 0.20 &  \cellcolor{tabfirst}0.908 &  \cellcolor{tabfirst}29.98 &                      0.253 & 0.35

    \end{tabular}
    }
\end{table*}


\begin{figure*}[tb]
	\centering
	\includegraphics[width=1\linewidth]{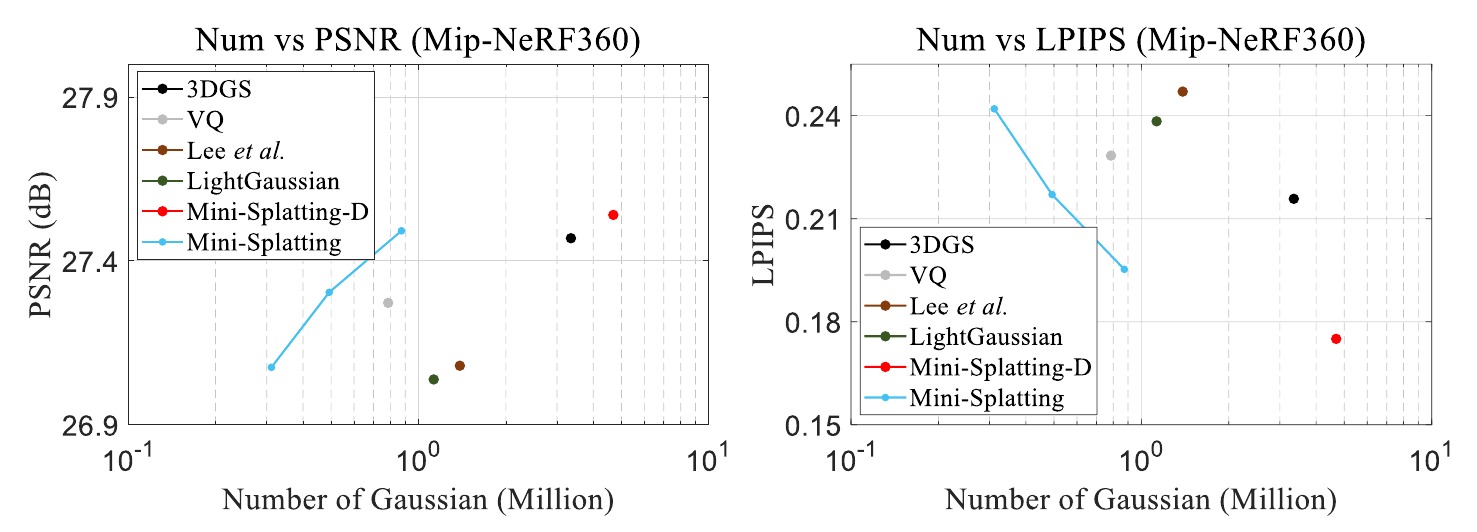}
    \vspace{-0.7cm}
    \caption{Number-Quality curves of our algorithms and other baselines on the Mip-NeRF360 dataset.}
	\label{fig:num_quality}
\end{figure*}

\begin{figure*}[tb]
	\centering
	\includegraphics[width=1\linewidth]{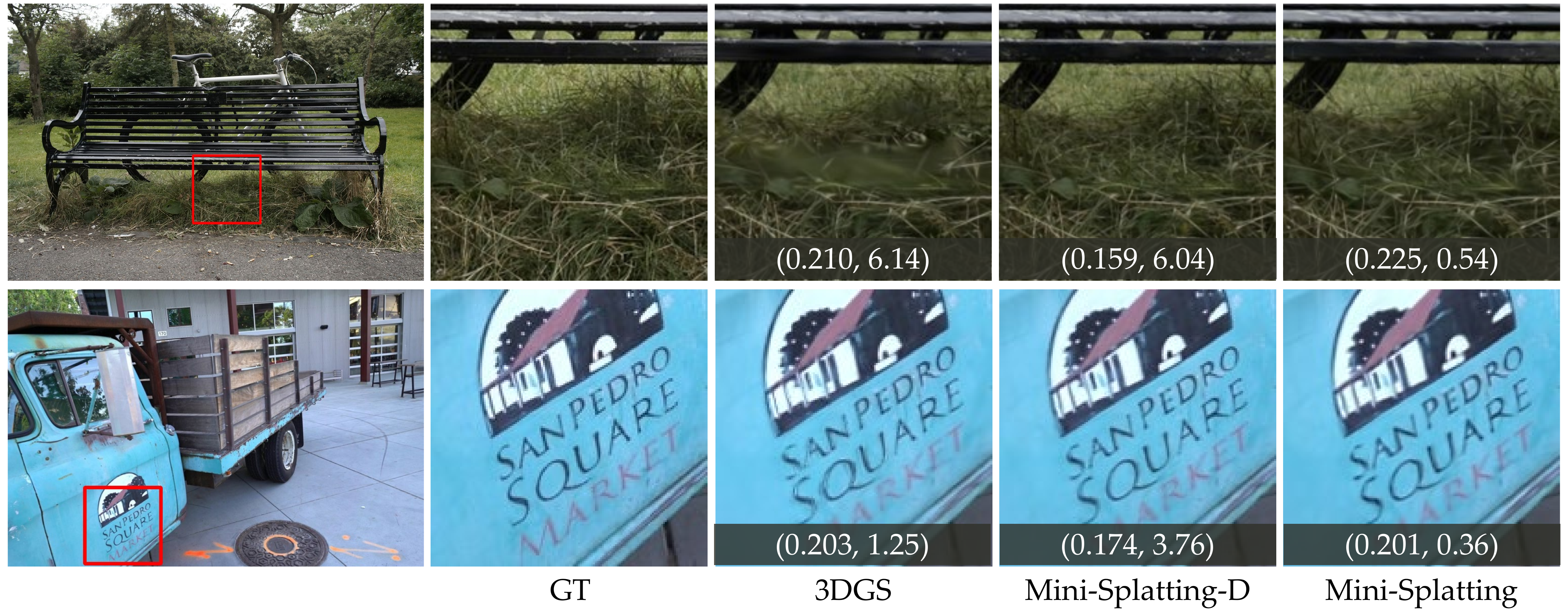}
    \vspace{-0.7cm}
    \caption{Qualitative results (\textit{bicycle} and \textit{truck}) of our Mini-Splatting and 3DGS. LPIPS and number of Gaussians in million are displayed as (LPIPS, Number).}
	\label{fig:qual}
\end{figure*}

\subsection{Experimental Results}

\noindent\textbf{Rendering Quality.} The quantitative results of various methods are presented in Table \ref{tab:quant}. We compare our Mini-Splatting and Mini-Splatting-D against the baseline 3DGS and other NeRF-based algorithms, utilizing standard metrics such as PSNR, SSIM, and LPIPS. The table demonstrates that, in comparison to the 3DGS baseline, our densification method, Mini-Splatting-D, exhibits superior performance across most metrics, highlighting the significant impact of Gaussian spatial distribution in rendering. Notably, our Mini-Splatting-D even surpasses the state-of-the-art neural rendering algorithm, Zip-NeRF \cite{barron2023zip}, in terms of SSIM and LPIPS on the Mip-NeRF360 dataset. Moreover, our Mini-Splatting is comparable to 3DGS, despite employing only a fraction ($7\times$ fewer) of the Gaussians. Our algorithms exhibit a decreased PSNR on the Tanks\&Temples dataset. This can be attributed to large sky areas in the scans (\ie, \textit{train}) within Tanks\&Temples, which our depth-based strategy does not accurately model. Further analysis is provided in the appendix.

As depicted in Fig. \ref{fig:num_quality}, we present the number-quality curves of our algorithms alongside 3DGS. The curve for our Mini-Splatting is generated by controlling the sampling ratio during the importance-weighted sampling process. Additionally, we compare against several recent Gaussian-Splatting-based algorithms, incorporating Gaussian pruning within their pipelines. Specifically, we apply the pruning technique proposed by \cite{li2023compressing} to 3DGS, denoted as VQ, and re-implement the prune and recovery stage outlined by \cite{fan2023lightgaussian} based on their official repository, referred to as LightGaussian. The results reported by Lee \etal \cite{lee2023compact} are collected from their paper. Notably, with a larger number of Gaussians, our Mini-Splatting-D demonstrates superior rendering quality, and it is clear that our Mini-Splatting outperforms all its counterparts at similar numbers of Gaussians.

The rendered images are presented in Fig. \ref{fig:qual}. These qualitative outcomes are in line with the quantitative results provided in Table \ref{tab:quant} and Fig. \ref{fig:num_quality}. Notably, our Mini-Splatting-D exhibits superior rendering quality in terms of finer details, while our Mini-Splatting maintains an acceptable rendering quality despite employing a limited number of Gaussians. These results indicate the effectiveness of our densification and simplification algorithm.

\noindent\textbf{Resource Consumption.} We also present resource consumption metrics, including training time, training peak GPU memory usage, rendering FPS, and rendering peak GPU memory usage, in Table \ref{tab:resource}, where peak memory consumption is measured using \texttt{torch.cuda.max\_memory\_allocated()}. In outdoor scenes, our Mini-Splatting-D, despite employing more Gaussians, exhibits similar training time and memory consumption to 3DGS due to our strategy of constraining SH coefficients during densification. Furthermore, our efficient representation significantly accelerates both training and rendering processes while reducing peak memory usage. In indoor scenes, both our Mini-Splatting-D and Mini-Splatting learn denser Gaussian distributions, yet the results still show an acceptable resource consumption of our algorithms. Additionally, we successfully train and evaluate our Mini-Splatting* on a low-cost graphics card (a GTX 1060 6G GPU). This highlights the potential for on-device neural rendering and reconstruction on consumer-grade machines.

\begin{table*}[t]
    \renewcommand{\tabcolsep}{1pt}
    \centering
    \caption{Resource consumption of our algorithms and 3DGS on Mip-NeRF360. Mini-Splatting* indicates Mini-Splatting trained on a GTX 1060 6G GPU, and all other models are trained with a RTX 3090 GPU. 
    }
    \label{tab:resource} 
    \resizebox{0.93\linewidth}{!}{
\begin{tabular}{@{}l@{\,\,}|c|cc|cc|c|cc|cc}
     & \multicolumn{5}{c|}{Mip-NeRF360 (Outdoor)} & \multicolumn{5}{c}{Mip-NeRF360 (Indoor)}\\ \hline
     \multirow{2}{*}{Method | Metric} & \multirow{2}{*}{Num} & \multicolumn{2}{c|}{Training} & \multicolumn{2}{c|}{Rendering} & \multirow{2}{*}{Num} & \multicolumn{2}{c|}{Training} & \multicolumn{2}{c}{Rendering} \\      
     &  & Time & Memory & FPS    & Memory &  & Time & Memory & FPS    & Memory  \\ \hline

    3DGS~\cite{kerbl20233d} & 4.86  & 30m8s & 7.45GB & 98    & 2.79GB & 1.46  & 24m41s & 2.75GB & 151   & 1.07GB \\ \hline
    Mini-Splatting-D & 5.40   & 31m48s & 7.45GB & 83    & 3.12GB & 3.80   & 40m13s & 5.55GB & 83    & 2.46GB \\
    Mini-Splatting & 0.57  & 17m56s & 2.61GB & 410   & 0.40GB & 0.40   & 27m2s & 2.77GB & 362   & 0.35GB \\
    Mini-Splatting* & 0.57  & 101m11s & 2.61GB & 64    & 0.40GB & 0.40   & 154m & 2.82GB & 40    & 0.35GB \\

    \end{tabular}%
    }
\end{table*}



\begin{figure*}[tb]
	\centering
	\includegraphics[width=0.8\linewidth]{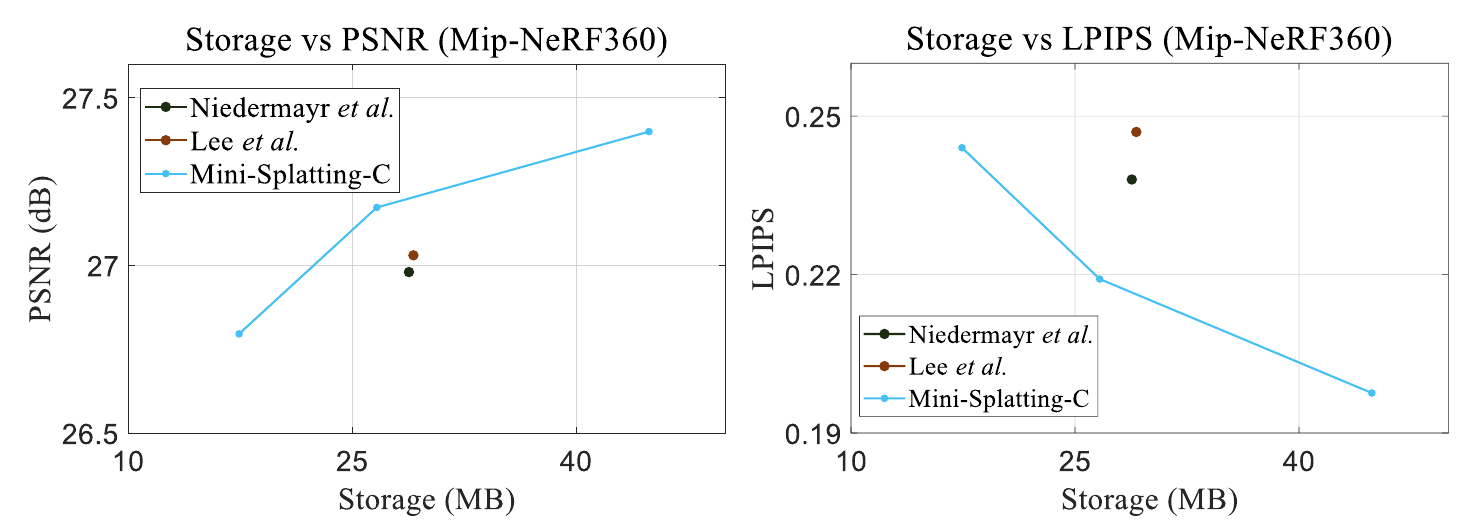}
    \vspace{-0.3cm}
    \caption{Rate-distribution curves for our Mini-Splatting-C and recent compression-oriented works \cite{niedermayr2023compressed, lee2023compact} applied to 3DGS.}
	\label{fig:storage}
\end{figure*}




\noindent\textbf{Storage Compression.} As depicted in Fig.  \ref{fig:storage}, we provide rate-distribution curves comparing our Mini-Splatting-C with two compression-oriented methods  \cite{niedermayr2023compressed, lee2023compact} applied to 3DGS. Results for these baseline approaches are collected from their respective papers. Due to the optimized and simplified Gaussian representation, our Mini-Splatting-C, which solely integrates basic post-processing techniques, demonstrates superior performance compared to other methods primarily tailored for storage compression.

\subsection{Ablation Studies}

\begin{table}[t]
\parbox{.5\linewidth}{
\centering
\caption{Ablation study of densification.}
\label{tab:abl_densification}
\resizebox{0.45\textwidth}{!}{%
      \begin{tabular}{@{}l@{\,\,}|cccc}

        \Xhline{2.0\arrayrulewidth}
      & SSIM $\uparrow$ & PSNR $\uparrow$ & LPIPS $\downarrow$ & Num \\ \hline
    Baseline & 0.815  & 27.47  & 0.216  & 3.35  \\
    + Blur Split & 0.819  & 27.47  & 0.195  & 3.74  \\
    + Depth Reinit & 0.832  & 27.54  & 0.175  & 4.32  \\
        \Xhline{2.0\arrayrulewidth}
    \end{tabular}%
}
}
\hfill
\parbox{.42\linewidth}{
\centering
\caption{Ablation study of depth reinitialization.}
\label{tab:abl_depth}
\resizebox{0.38\textwidth}{!}{%
      \begin{tabular}{@{}l@{\,\,}|cccc}
        \Xhline{2.0\arrayrulewidth}
      & SSIM $\uparrow$ & PSNR $\uparrow$ & LPIPS $\downarrow$ & Num \\ \hline
    Blending & 0.513  & 17.67  & 0.475  & 4.01  \\
    Center & 0.832  & 27.57  & 0.176  & 4.70  \\
    Mid   & 0.832  & 27.54  & 0.175  & 4.69  \\
        \Xhline{2.0\arrayrulewidth}
    \end{tabular}%
}%
}
\end{table}

\noindent\textbf{Densification.} We begin by presenting the ablation study of our densification in Table \ref{tab:abl_densification}. Starting from the 3DGS baseline, we incrementally introduce blur splitting and depth reinitialization steps. The results demonstrate a consistent improvement in rendering quality with the increasing number of Gaussians. Notably, we observe a high correlation between the number of Gaussians and the LPIPS metric, indicating that a dense and uniform spatial distribution of Gaussians positively influences human perception metrics.

We extend our analysis by comparing blending depth with depth obtained from Gaussian center and mid-point in Table \ref{tab:abl_depth}. As discussed in Sec. \ref{sec:Densification}, reinitializing Gaussians with blending depth may degrade rendering quality due to the presence of noise points resulting from depth artifacts. Although depth reinitialization using Gaussian center yields results comparable to those obtained from the mid-point, depth derived from the mid-point exhibits superior dense point cloud reconstruction, as illustrated in Fig. \ref{fig:merged_depth}, and holds potential for extension to normal estimation tasks.

\setlength{\columnsep}{10pt}
\begin{wrapfigure}[12]{R}{0.46\textwidth}
\vspace{-0.3cm}
\centering
\raisebox{5pt}[\dimexpr\height-1\baselineskip\relax]{
\includegraphics[width=1\linewidth]{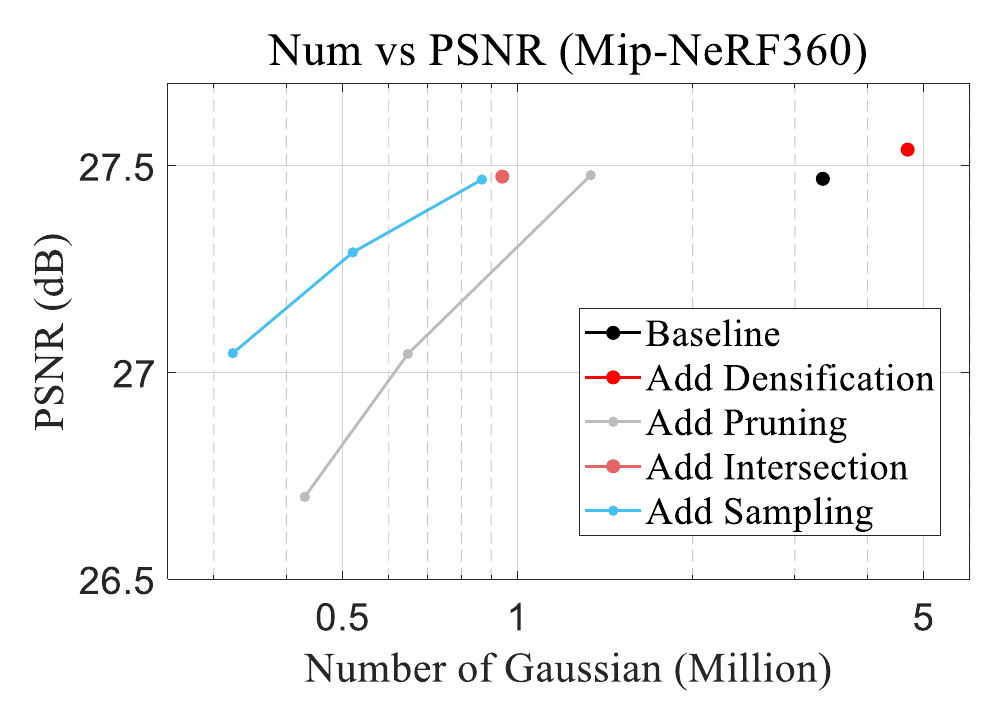}}
\vspace{-0.7cm}
\caption{Ablations of simplification.}
\label{fig:abl_simplification}
\end{wrapfigure} 
\noindent\textbf{Simplification.} The ablations of simplification on Mip-NeRF360 are depicted in Fig. \ref{fig:abl_simplification}. Starting from the 3DGS baseline, we first add densification to obtain our Mini-Splatting-D. Subsequently, we integrate direct pruning into Mini-Splatting-D and tune the pruning ratio to establish the baseline curve, Add Pruning, for our simplification algorithm. It is worth noting that this baseline encompasses the combination of Mini-Splatting-D and direct pruning, surpassing 3DGS with pruning (\ie, VQ) as depicted in Fig. \ref{fig:num_quality}. To underscore the effectiveness of our intersection preserving and sampling techniques, we add intersection preserving and then sampling to Mini-Splatting-D (Add Intersection and Add Sampling). The results clearly demonstrate that our simplification algorithm outperforms direct pruning, achieving approximately a halved reduction in the number of Gaussians while maintaining comparable rendering quality. This result indicates the effectiveness of the proposed simplification method.


	




\section{Conclusion}

We introduced Mini-Splatting, which aimed at representing scene with a constrained nubmer of Gaussians, through our proposed densification and simplification algorithm. Our Gaussian densification process, involving blur splitting and depth initialization, facilitates the construction of a dense Gaussian representation, while our simplification algorithm, featuring intersection preserving and Gaussian sampling, effectively limits the number of Gaussians while preserving high-quality rendering. Our experimental results show the effectiveness of Mini-Splatting and its variants across benchmarks, encompassing not only rendering quality but also resource consumption and storage compression.


\smallskip\noindent\textbf{Acknowledgements.} This work was jointly supported by the Young Scientists Fund of the National Natural Science Foundation of China (42301520), the Research Grants Council of Hong Kong (25206524), the Platform Project of Unmanned Autonomous Systems Research Centre (P0049516), the Seed Project  of Smart Cities Research Institute (P0051028), the Research Team Cultivation Program of ShenZhen University (2023JCT003), and the Pearl River Talent Program (2021JC02G046).

\clearpage  

\newpage
\appendix

\section*{Appendix}

\section{Visualization of Elliptical Gaussians}

\begin{figure*}[h]
	\centering
	\includegraphics[width=1\linewidth]{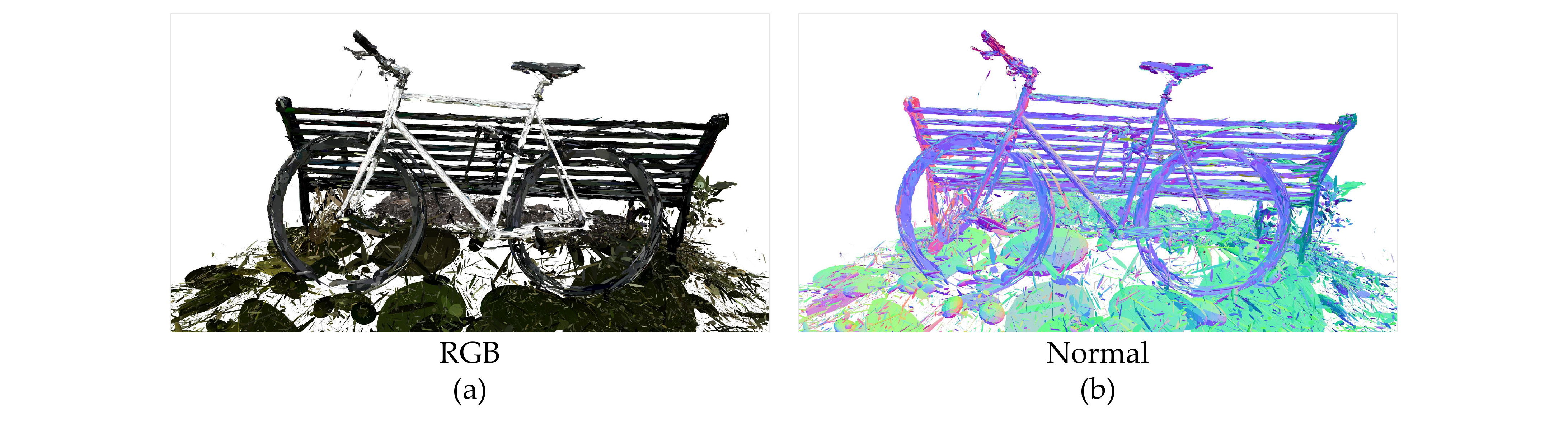}
    \vspace{-0.7cm}
    \caption{Visualization of Elliptical Gaussians. (a): RGB colors derived from the first order of SH coefficients. (b): Normals of the corresponding ellipsoids. Gaussians with an opacity lower than 0.1 have been removed for better visualization.}
	\label{fig:basic_as}
\end{figure*}

We visualize 3D Gaussians as ellipsoids using Open3D, as illustrated in Fig. \ref{fig:basic_as}. Within the Gaussian representation, the Gaussian centers, akin to point clouds, can be interpreted as the geometric structure (see Fig. \ref{fig:basic_as} (a)). The elliptical Gaussians are distributed around the geometric surface, resembling surface segments with consistent normal vectors, as depicted in Fig. \ref{fig:basic_as} (b). This inherent geometric structure of the Gaussian representation supports our analysis from the perspective of point clouds.

\section{Gaussian Splatting with Dense Initialization}


We additionally offer an analysis of 3DGS using dense initialization. In this experiment, we acquire dense point clouds through Multi-View Stereo (MVS) \cite{schoenberger2016mvs}, and substitute these dense points with the original sparse points for initialization purposes. To prevent memory overflow, dense point clouds from each scene are randomly sampled to 2 million points.

\begin{figure*}[h]
	\centering
	\includegraphics[width=1\linewidth]{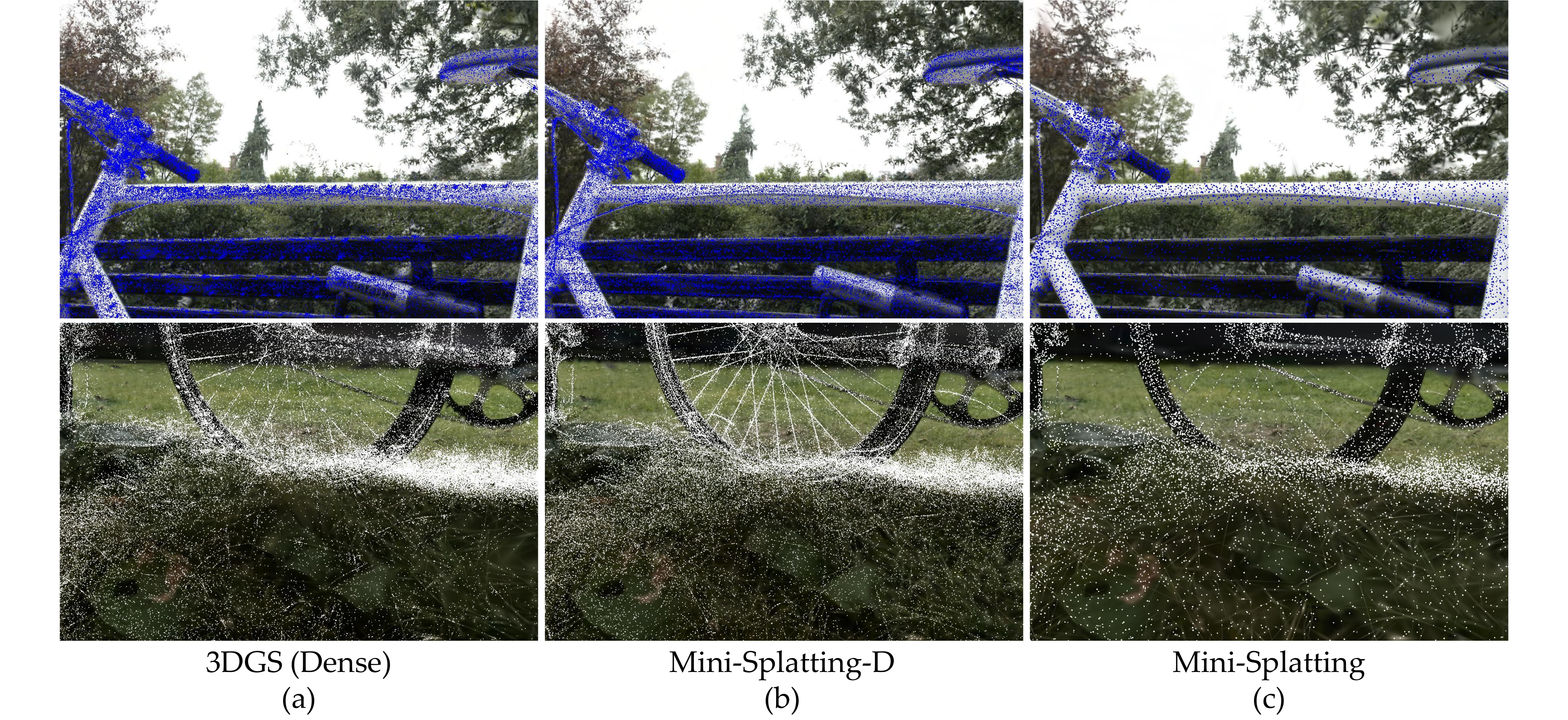}
    \vspace{-0.7cm}
    \caption{Visual analysis of Gaussian centers. (a): Projected Gaussian centers of the densely initialized 3DGS model. (b) and (c): Projected Gaussian centers of our Mini-Splatting-D and Mini-Splatting, both initialized with sparse points.}
    \label{fig:dense_center}
\end{figure*}

As depicted in Fig. \ref{fig:dense_center}, we project the Gaussian centers of densely initialized 3DGS, Mini-Splatting-D, and Mini-Splatting onto the rendered image, represented as blue and white points. In comparison with Mini-Splatting-D and Mini-Splatting, 3DGS (Dense) still exhibits the issue of `overlapping'. Specifically, Gaussian centers tend to cluster around the top tube and spokes of the bicycle, while our Mini-Splatting-D and Mini-Splatting demonstrate a more uniform point distribution. Regarding the phenomenon of `under-reconstruction' (\ie, the second row), dense initialization can mitigate this issue, yet our Mini-Splatting-D still showcases denser points in the corresponding area.

We further present quantitative results of our method alongside 3DGS with sparse and dense initialization in \ref{tab:dense_init}. We can see a significant improvement in all methods with dense initialization. Remarkably, all methods demonstrate significant enhancements with dense initialization. This outcome underscores the potential for further refinement of our algorithm through the integration of stereo matching techniques. It is noteworthy that our Mini-Splatting-D (Sparse) continues to exhibit superior SSIM and LPIPS scores compared to 3DGS (Dense), underscoring the effectiveness of our densification algorithm.

\begin{table*}[hh]
    \renewcommand{\tabcolsep}{1pt}
    \centering
    \caption{Quantitative evaluation of our Mini-Splatting and 3DGS with sparse and dense initialization on the Mip-NeRF360 dataset.}
    \label{tab:dense_init}    
    \resizebox{0.85\linewidth}{!}{
    \begin{tabular}{@{}l@{\,\,}|cccc|cccc}

    \Xhline{2.0\arrayrulewidth}
    Dataset & \multicolumn{4}{c|}{Mip-NeRF360 (Sparse)} & \multicolumn{4}{c}{Mip-NeRF360 (Dense)} \\
    
  Method $|$ Metric & SSIM $\uparrow$ & PSNR $\uparrow$ & LPIPS $\downarrow$ & Num   & SSIM $\uparrow$ & PSNR $\uparrow$ & LPIPS $\downarrow$ & Num   \\\hline

    3DGS  & 0.815  & 27.47  & 0.216  & 3.35  & 0.831  & 27.78  & 0.180  & 4.40  \\ \hline
    Mini-Splatting-D & 0.832  & 27.54  & 0.175  & 4.69  & 0.838  & 27.76  & 0.171  & 4.71  \\
    Mini-Splatting & 0.822  & 27.22  & 0.218  & 0.48  & 0.825  & 27.31  & 0.215  & 0.48  \\
    \Xhline{2.0\arrayrulewidth}
    \end{tabular}
    }
\end{table*}

\section{Artifacts of Blending Depth}

\begin{figure*}[h]
	\centering
	\includegraphics[width=1\linewidth]{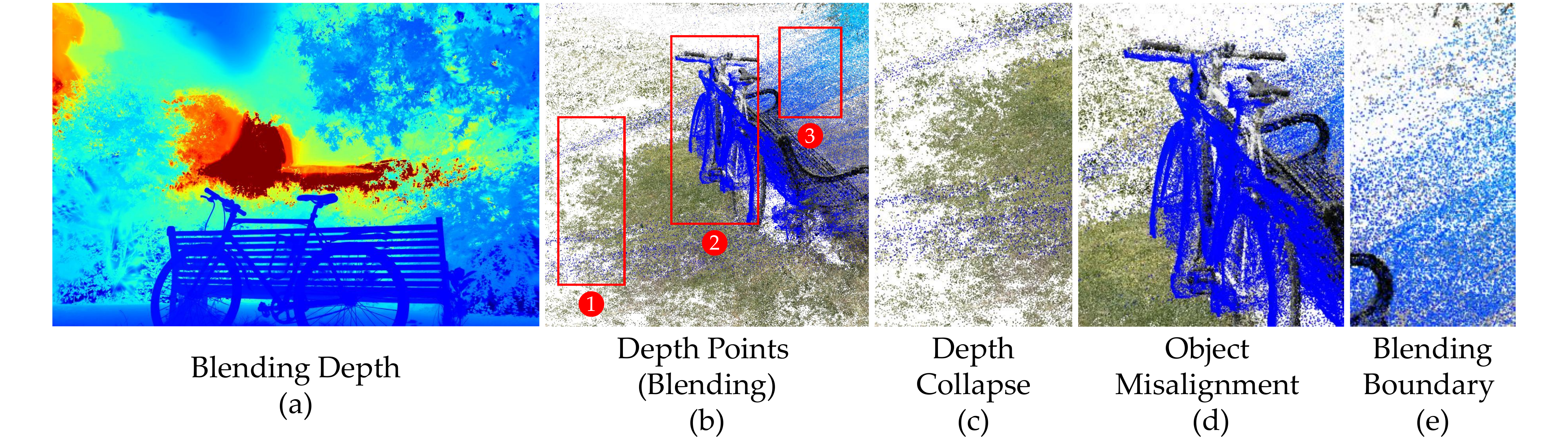}
    \vspace{-0.7cm}
    \caption{Visual analysis of blending depth and depth points from 3DGS \cite{kerbl20233d}. (a): Rendered depth map from alpha blending. (b): Reconstructed depth points from alpha blending, with red boxes highlighting three significant artifacts, (c): depth collapse, (d): object misalignment, and (e): blending boundary.}
    \label{fig:blending_depth}
\end{figure*}

We display rendered depth obtained from alpha blending in Fig. \ref{fig:blending_depth} (a), and subsequently reproject the rendered depth into world space in Fig. \ref{fig:blending_depth} (b). Artifacts, including depth collapse, object misalignment and blending boundary, are emphasized in Figs. \ref{fig:blending_depth} (c), (d) and (e). These failures can primarily be attributed to the alpha blending of multiple Gaussians.

\textbf{Depth Collapse.} Several segments of the background scene exhibit unreasonably low depth values. This issue arises primarily because 3DGS employs a default background color (\ie, black for the Mip-NeRF360 dataset), which is incorporated into the blending color to generate the final output. For black objects situated in the background, 3DGS optimizes these segments based on the default background color, resulting in a failure to accurately learn the geometry. Consequently, we observe a relatively low accumulated opacity value along the ray, leading to a tendency for the corresponding depth to collapse into a low depth value.

\textbf{Object Misalignment.} It is clear that the depth points do not align with the Gaussian centers in Fig. \ref{fig:blending_depth} (d). This discrepancy arises due to the presence of large Gaussians around the object and floaters positioned in front of the camera. These Gaussians, optimized to represent reflection and noise, result in inaccurate depth estimation through alpha blending.

\textbf{Blending Boundary.} Recovering a sharp boundary poses a challenge for alpha blending due to the formulation of the weighted sum. Along the boundary, all Gaussians exhibit relatively low blending weights, resulting in a smooth image edge in the depth map. Consequently, the issue of blending boundary becomes apparent in the world space.

\section{Calculation of Gaussian Depth}

Given an elliptical Gaussian defined by its scale $\bs=(s_x, s_y, s_z)$, we model the Gaussian as an ellipsoid in 3D space as $g(x, y, z)=\frac{x^2}{s_x^2} + \frac{y^2}{s_y^2} + \frac{z^2}{s_z^2} = 1$. For an input ray $\br(t)=\bo+t\bd$, where $\bo=(o_x, o_y, o_z)$ and $\bd=(d_x, d_y, d_z)$ are the ray origin and direction transformed into the coordinate system of the ellipsoid, we can compute the intersection points of the ray and the ellipsoid by solving for the parameter $t$ that satisfies the equation.

First, substitute the parametric representation of the ray into the equation of the ellipsoid as follows:
\begin{equation}
    \frac{(o_x+t d_x)^2}{s_x^2} + \frac{(o_y+t  d_y)^2}{s_y^2} + \frac{(o_z+t d_z)^2}{s_z^2} = 1.
\end{equation}
Then, expand and rewrite the equation as:
\begin{equation}
    (\frac{d_x^2}{s_x^2} + \frac{d_y^2}{s_y^2} + \frac{d_z^2}{s_z^2})t^2 + 2\left(\frac{o_x d_x}{s_x^2} + \frac{o_y d_y}{s_y^2} + \frac{o_z d_z}{s_z^2}\right)t + \left(\frac{o_x^2}{s_x^2} + \frac{o_y^2}{s_y^2} + \frac{o_z^2}{s_z^2} - 1\right) = 0.
\end{equation}
This equation is quadratic in $t$, and we can solve it using the quadratic formula:
\begin{equation}
    t = \frac{-b \pm \sqrt{b^2 - 4ac}}{2a}, 
\end{equation}
where $a = \frac{d_x^2}{s_x^2} + \frac{d_y^2}{s_y^2} + \frac{d_z^2}{s_z^2}$, $b = 2\left(\frac{o_x d_x}{s_x^2} + \frac{o_y d_y}{s_y^2} + \frac{o_z d_z}{s_z^2}\right)$, and $c = \frac{o_x^2}{s_x^2} + \frac{o_y^2}{s_y^2} + \frac{o_z^2}{s_z^2} - 1$. 
Thus, the position of the mid-point $\bp^{mid}$ can be calculated as
\begin{equation}
    \bp^{mid}=\br(t^{mid})=\br(-\frac{b}{2a})=\bo+-\frac{b}{2a}\bd, \quad \text{where } \Delta=b^2 - 4ac.
\end{equation}
The discriminant $\Delta$ is used to determine whether real intersection points exist, and our Gaussian depth $d^{mid}$ can be also obtained as $t^{mid}$.

Another intuitive formulation involves calculating the optimized $t$ where the maximum density of the Gaussian distribution along the ray is reached. Given a Gaussian distribution defined by its scale $\bs=(s_x, s_y, s_z)$, the distribution can be formulated as:
\begin{equation}
    \mathcal{N}(\mathbf{x}|\boldsymbol{0},\boldsymbol{\Sigma}) = \frac{1}{(2\pi)^\frac{n}{2}|\boldsymbol{\Sigma}|^\frac{1}{2}} \exp\left(-\frac{1}{2}\mathbf{x}^\top \boldsymbol{\Sigma}^{-1}\mathbf{x}\right), \quad \text{where } \boldsymbol{\Sigma}=diag(s_x^2, s_y^2, s_z^2).
\end{equation}
With the input ray $\br(t)=\bo+t\bd$, we can calculate the optimized $t^{opt}$.

First, the substitution and simplification:
\begin{equation}
\begin{split}
    \mathcal{N}(\mathbf{x}|\boldsymbol{0},\boldsymbol{\Sigma}) &= \frac{1}{(2\pi)^\frac{n}{2}|\boldsymbol{\Sigma}|^\frac{1}{2}} \exp\left(-\frac{1}{2} \left(\frac{o_x^2}{s_x^2} + \frac{o_y^2}{s_y^2} + \frac{o_z^2}{s_z^2}\right) \right. \\
&\left. -\frac{t}{2} \left(\frac{2o_xd_x}{s_x^2} + \frac{2o_yd_y}{s_y^2} + \frac{2o_zd_z}{s_z^2}\right) - \frac{t^2}{2} \left(\frac{d_x^2}{s_x^2} + \frac{d_y^2}{s_y^2} + \frac{d_z^2}{s_z^2}\right)\right)
\end{split}
\end{equation}
Denote the equation as:
\begin{equation}
\mathcal{N}(\mathbf{x}|\boldsymbol{0},\boldsymbol{\Sigma}) = A \exp(Bt + Ct^2),
\end{equation}
with $A = \frac{1}{(2\pi)^\frac{n}{2}|\boldsymbol{\Sigma}|^\frac{1}{2}} \exp\left(-\frac{1}{2} \left(\frac{o_x^2}{s_x^2} + \frac{o_y^2}{s_y^2} + \frac{o_z^2}{s_z^2}\right)\right)$, $B = -\frac{1}{2} \left(\frac{2o_xd_x}{s_x^2} + \frac{2o_yd_y}{s_y^2} + \frac{2o_zd_z}{s_z^2}\right)$ and $C = -\frac{1}{2} \left(\frac{d_x^2}{s_x^2} + \frac{d_y^2}{s_y^2} + \frac{d_z^2}{s_z^2}\right)$. To find the maximum, we take the derivative with respect to $t$ and set it to zero:
\begin{equation}
\begin{split}
\frac{d}{dt} (A \exp(Bt + Ct^2)) = 0, \\
A \exp(Bt + Ct^2) \left(2Ct+B \right) = 0
\end{split}
\end{equation}
We can ignore \( A \exp(Bt + Ct^2) \) because it will never be zero, so we can obtain $t^{opt}=-\frac{B}{2C}$.

Note that we have $t^{mid}=t^{opt}$. We opt for the former formulation because both yield the same numerical result, yet the former one offers an additional discriminant. This discriminant aids in distinguishing whether an intersection point of the ray and the ellipsoid exists.

\section{Importance Metric}

In our experiments, we observe that utilizing blending weight-based importance yields relatively acceptable performance in pruning. Specifically, as depicted in Fig. 7 of the main paper, direct pruning on 3DGS using the accumulated blending weights as the importance metric (\ie, VQ \cite{li2023compressing}) can achieve superior results compared to several recent works \cite{fan2023lightgaussian, lee2023compact}. Thus, we further construct our importance metric based on blending weight for our importance-weighted sampling approach.

Here, we assign the the aforementioned importance as $I_i^1=\sum_{j=1}^K w_{ij}$, where $w_{ij}$ represents the blending weight of the Gaussian $\cG_i$ with the $j$th ray, and $K$ stands for the total count of rays intersecting with $\cG_i$. It is noteworthy that this importance $I_i^1$ demonstrates superior performance within indoor settings, characterized by Gaussians exhibiting low uncertainty (\ie, not a floater) and sharing similar scales. However, outdoor scans often feature additional Gaussians representing background elements such as the sky or distant objects. These Gaussians tend to be preserved when solely relying on blending weights. Thus, for each image in outdoor scene, we only accumulate the blending weights of Gaussians which provide intersection points (mid-points) with rays, and also include the projected area to limit the scale of preserved Gaussians. For the $m$th image, we denote the  importance $I_i^{(m)}$, and then aggregate all $I_i^{(m)}$ as our second importance metric $I_i^2$ for $\cG_i$ as:
\begin{equation}
I_i^2=\sum_{m=1}^M I_i^{(m)} \cdot \delta(i \in \mathcal{I}_{max}^{(m)}), \quad I_i^{(m)}=\sum_{j=1}^K \frac{w_{ij}^{(m)}}{S_{i}^{(m)}}.
\end{equation}
Here, $\delta$ represents the indicator function, and $\mathcal{I}_{max}^{(m)}$ denotes the collection of all indexes in the $m$th rendered indexes $i_{max}^{(m)}(\bx)$. $S_{i}^{(m)}$ is the projected area of $\cG_i$ on the $m$th image.

\begin{figure*}[h]
	\centering
	\includegraphics[width=0.9\linewidth]{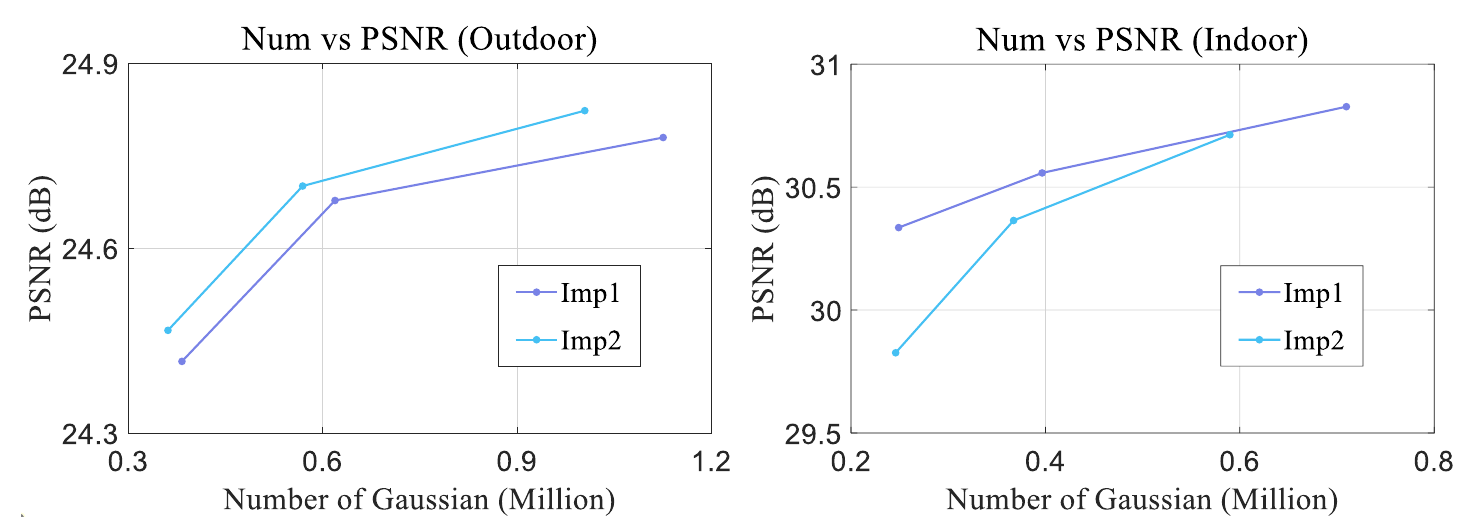}
    \vspace{-0.3cm}
    \caption{Comparison between the importance metrics. We denote $I_i^1$ as Imp1 and $I_i^2$ as Imp2.}
	\label{fig:imp}
\end{figure*}

We further compare two importance metrics as Imp1 $I_i^1$ and Imp2 $I_i^2$ on the Mip-NeRF360 dataset, as depicted in Fig. \ref{fig:imp}. The results indicate that combinations of blending weight and other information exhibit varying performance on indoor and outdoor scans. We posit that the design of importance metrics is case-dependent and hand-crafted. Thus, this part is presented as an experimental trick in the appendix.

\section{Details of Mini-Splatting and its Variants}

\begin{algorithm}[h]
\caption{Densification and Simplification}
\label{ms}
\begin{algorithmic}[1]
\State Gaussians $\gets$ SfM Points \Comment{Initialization}
\State $i \gets 0$ \Comment{Iteration Count}
\While{not converged}
    \State Optimization()
    \If{$i < $ DensificationIteration}
        \If{IsRefinementIteration($i$)}
           \State BlurSplit() \Comment{Our Blur Split}
           \State SplitAndClone() \Comment{Split and Clone of 3DGS}
        \EndIf
        \If{IsReinitIteration($i$)}
           \State DepthReinit() \Comment{Our Depth Reinitialization}
        \EndIf        
    \Else
        \If{$i=$ SimplificationIteration1}
            \State Intersection() \Comment{Our Intersection Preserving}
            \State Sampling() \Comment{Our Importance-weighted Sampling}
            \State Gaussians $\gets$ Gaussian Centers \Comment{Initialization}
        \EndIf     
        \If{$i=$ SimplificationIteration2}
            \State Intersection() \Comment{Our Intersection Preserving}
            \State Pruning()  \Comment{Directly Prune a Few Gaussians}
        \EndIf      
    \EndIf
    \State $i \gets i+1$
\EndWhile
\end{algorithmic}
\end{algorithm}

The pipeline of our Mini-Splatting is summarized in Algorithm \ref{ms}. The DensificationIteration is set as $15K$, and our depth reinitialization is enabled every $5K$ iterations. The SimplificationIteration1 and SimplificationIteration2 are set as $15K$ and $20K$, respectively. For Mini-Splatting-D, we simply disable the process of simplification.

The pipeline of Mini-Splatting-C is directly based on our Mini-Splatting. Given a pre-trained Mini-Splatting model, we store the Gaussian centers as \texttt{float32}, while other attributes, such as scales and SH coefficients, are processed using Region-adaptive Hierarchical Transform \cite{de2016compression} with the depth level set as 16 and the quantization step set as 0.02. The transformed coefficients are further losslessly compressed by the standard zip compression provided by NumPy (\ie, \texttt{numpy.savez\_compressed()}).

\section{Limitation and Future Work}

\begin{figure*}[h]
	\centering
	\includegraphics[width=1\linewidth]{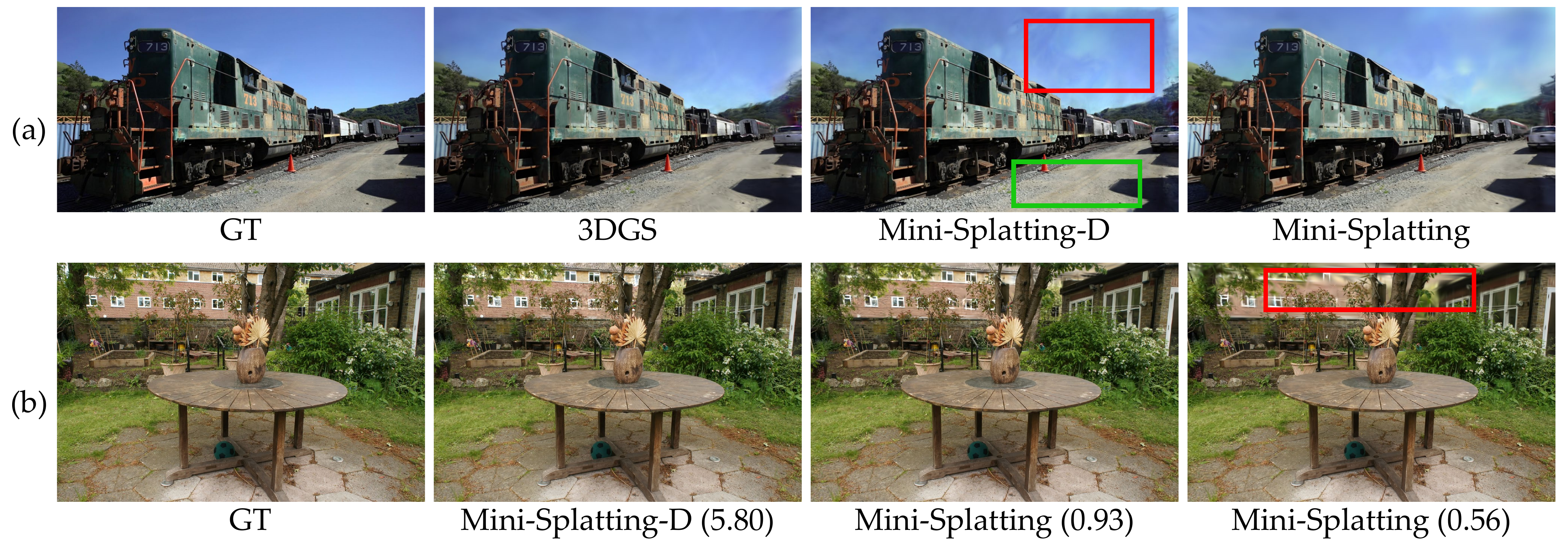}
    \vspace{-0.7cm}
    \caption{Limitations. (a): Our depth-based strategy fails in the area of sky. (b): The background objects exhibit significant distribution with a small number of Gaussians.}
	\label{fig:Limitations}
\end{figure*}

Our method employs a depth-based strategy to reinitialize Gaussian models. This strategy fails in areas without a certain depth value, such as the sky in \textit{train}, as shown in Fig. \ref{fig:Limitations} (a). This issue could potentially be addressed by considering multiview consistency and background removal. Moreover, the number of Gaussians in our Mini-Splatting is manually controlled by the sampling ratio. However, finding the minimal number of Gaussians while maintaining high-quality rendering remains a challenge. As illustrated in Fig. \ref{fig:Limitations} (b), a high sampling ratio leads to distortion in the background. This problem may be alleviated by considering uncertainty for each image, which we leave for future study.

\section{Additional Quantitative Results}

We provide additional quantitative results on all three datasets in Table \ref{tab:quant_SSIM}. Our algorithm demonstrates a consistent improvement compared to 3DGS in most scans and even surpasses Zip-NeRF in terms of SSIM and LPIPS metrics.

\begin{table*}[tb]
    \renewcommand{\tabcolsep}{1pt}
    \centering
    \caption{Quantitative evaluation of our Mini-Splatting and previous works. 3DGS* indicates the retrained model from the official implementation.}
    \label{tab:quant_SSIM}   
    \resizebox{1\linewidth}{!}{
    \begin{tabular}{l|ccccc|cccc|cc|cc}

    & \multicolumn{13}{c}{\textbf{SSIM}} \\

      & \textit{bicycle} & \textit{flowers} & \textit{garden} & \textit{stump} & \textit{treehill} & \textit{room} & \textit{counter} & \textit{kitchen} & \textit{bonsai} & \textit{truck} & \textit{train} & \textit{johnson} & \textit{playroom} \\ \hline

mip-NeRF 360~\cite{barron2022mipnerf360} &    
                  0.693 &
 0.583 &                      0.816 &
             0.746 &                      0.632 &                      0.913 &
        0.895 &                      0.920 &  
                    0.939 &
   0.857 &                      0.660 &  \cellcolor{tabthird}0.901 &                      0.900 \\
Zip-NeRF~\cite{barron2023zip}            &    
                  0.769 &  \cellcolor{tabfirst}0.642 &                      0.860 &  \cellcolor{tabthird}0.800 &  \cellcolor{tabfirst}0.681 & \cellcolor{tabsecond}0.925 &
        0.902 & \cellcolor{tabsecond}0.928 &  
\cellcolor{tabfirst}0.949 &
    -  &                       -  &
            -  &                       -  \\  
3DGS~\cite{kerbl20233d}                  &  \cellcolor{tabthird}0.771 &  \cellcolor{tabthird}0.605 & \cellcolor{tabsecond}0.868 &
             0.775 &                      0.638 &                      0.914 &  \cellcolor{tabthird}0.905 &                      0.922 &  
                    0.938 &  \cellcolor{tabthird}0.879 &  \cellcolor{tabthird}0.802 &       
               0.899 &                      0.906 \\
3DGS~\cite{kerbl20233d}*                 &    
                  0.764 &  \cellcolor{tabthird}0.605 &  \cellcolor{tabthird}0.867 &
             0.772 &                      0.632 &                      0.919 & \cellcolor{tabsecond}0.909 &  \cellcolor{tabthird}0.927 &  
\cellcolor{tabthird}0.942 & \cellcolor{tabsecond}0.882 & \cellcolor{tabsecond}0.814 &       
               0.900 &  \cellcolor{tabthird}0.907 \\ \hline
Mini-Splatting-D                         &  \cellcolor{tabfirst}0.798 &  \cellcolor{tabfirst}0.642 &  \cellcolor{tabfirst}0.878 & \cellcolor{tabsecond}0.804 &  \cellcolor{tabthird}0.640 &  \cellcolor{tabfirst}0.928 &  \cellcolor{tabfirst}0.913 &  \cellcolor{tabfirst}0.934 & \cellcolor{tabsecond}0.946 &  \cellcolor{tabfirst}0.890 &  \cellcolor{tabfirst}0.817 &  \cellcolor{tabfirst}0.905 & \cellcolor{tabsecond}0.908 \\
Mini-Splatting                           & \cellcolor{tabsecond}0.772 & \cellcolor{tabsecond}0.626 &                      0.847 &  \cellcolor{tabfirst}0.806 & \cellcolor{tabsecond}0.654 &  \cellcolor{tabthird}0.921 &
        0.904 &                      0.926 &  
                    0.940 &
   0.874 &                      0.796 & \cellcolor{tabsecond}0.904 &  \cellcolor{tabfirst}0.912 \\

    \multicolumn{14}{c}{} \\
    & \multicolumn{13}{c}{\textbf{PSNR}} \\

      & \textit{bicycle} & \textit{flowers} & \textit{garden} & \textit{stump} & \textit{treehill} & \textit{room} & \textit{counter} & \textit{kitchen} & \textit{bonsai} & \textit{truck} & \textit{train} & \textit{johnson} & \textit{playroom} \\ \hline

mip-NeRF 360~\cite{barron2022mipnerf360} &    
                  24.40 & \cellcolor{tabsecond}21.64 &                      26.94 &
             26.36 & \cellcolor{tabsecond}22.81 &                      31.40 &  \cellcolor{tabfirst}29.44 & \cellcolor{tabsecond}32.02 & \cellcolor{tabsecond}33.11 &
   24.91 &                      19.52 &  \cellcolor{tabthird}29.14 &                      29.66 \\
Zip-NeRF~\cite{barron2023zip}            &  \cellcolor{tabfirst}25.80 &  \cellcolor{tabfirst}22.40 &  \cellcolor{tabfirst}28.20 &  \cellcolor{tabfirst}27.55 &  \cellcolor{tabfirst}23.89 &  \cellcolor{tabfirst}32.65 & \cellcolor{tabsecond}29.38 &  \cellcolor{tabfirst}32.50 &  
\cellcolor{tabfirst}34.46 &
    -  &                       -  &
            -  &                       -  \\  
3DGS~\cite{kerbl20233d}                  &  \cellcolor{tabthird}25.25 &
 21.52 &  \cellcolor{tabthird}27.41 &
             26.55 &                      22.49 &                      30.63 &
        28.70 &                      30.32 &  
                    31.98 & \cellcolor{tabsecond}25.19 &  \cellcolor{tabthird}21.10 &       
               28.77 &  \cellcolor{tabthird}30.04 \\
3DGS~\cite{kerbl20233d}*                 &    
                  25.19 &
 21.55 &  \cellcolor{tabthird}27.41 &
             26.59 &                      22.47 & \cellcolor{tabsecond}31.46 &  \cellcolor{tabthird}29.03 &                      31.27 &  
\cellcolor{tabthird}32.24 &  \cellcolor{tabfirst}25.43 &  \cellcolor{tabfirst}21.90 &       
               29.12 &                      29.95 \\ \hline
Mini-Splatting-D                         & \cellcolor{tabsecond}25.55 &
 21.50 & \cellcolor{tabsecond}27.67 &  \cellcolor{tabthird}27.11 &                      22.13 &  \cellcolor{tabthird}31.41 &
        28.72 &  \cellcolor{tabthird}31.75 &  
                    31.72 &  \cellcolor{tabfirst}25.43 &                      21.04 & \cellcolor{tabsecond}29.32 & \cellcolor{tabsecond}30.43 \\
Mini-Splatting                           &    
                  25.15 &  \cellcolor{tabthird}21.58 &                      26.84 & \cellcolor{tabsecond}27.26 &  \cellcolor{tabthird}22.72 &                      31.39 &
        28.41 &                      31.17 &  
                    31.57 &  \cellcolor{tabthird}25.09 & \cellcolor{tabsecond}21.28 &  \cellcolor{tabfirst}29.47 &  \cellcolor{tabfirst}30.49 \\

    \multicolumn{14}{c}{} \\
    & \multicolumn{13}{c}{\textbf{LPIPS}} \\

      & \textit{bicycle} & \textit{flowers} & \textit{garden} & \textit{stump} & \textit{treehill} & \textit{room} & \textit{counter} & \textit{kitchen} & \textit{bonsai} & \textit{truck} & \textit{train} & \textit{johnson} & \textit{playroom} \\ \hline

mip-NeRF 360~\cite{barron2022mipnerf360} &    
                  0.289 &
 0.345 &                      0.164 &
             0.254 &                      0.338 &  \cellcolor{tabthird}0.211 &
        0.203 &  \cellcolor{tabthird}0.126 &  
\cellcolor{tabthird}0.177 &
   0.159 &                      0.354 & \cellcolor{tabsecond}0.237 &                      0.252 \\
Zip-NeRF~\cite{barron2023zip}            &  \cellcolor{tabthird}0.208 & \cellcolor{tabsecond}0.273 &                      0.118 & \cellcolor{tabsecond}0.193 &  \cellcolor{tabfirst}0.242 & \cellcolor{tabsecond}0.196 & \cellcolor{tabsecond}0.185 & \cellcolor{tabsecond}0.116 &  
\cellcolor{tabfirst}0.173 &
    -  &                       -  &
            -  &                       -  \\  
3DGS~\cite{kerbl20233d}                  & \cellcolor{tabsecond}0.205 &
 0.336 & \cellcolor{tabsecond}0.103 &
             0.210 &                      0.317 &                      0.220 &
        0.204 &                      0.129 &  
                    0.205 &  \cellcolor{tabthird}0.148 &  \cellcolor{tabthird}0.218 &  \cellcolor{tabthird}0.244 & \cellcolor{tabsecond}0.241 \\
3DGS~\cite{kerbl20233d}*                 &    
                  0.210 &
 0.336 &  \cellcolor{tabthird}0.107 &
             0.217 &                      0.325 &                      0.219 &
        0.200 &  \cellcolor{tabthird}0.126 &  
                    0.203 & \cellcolor{tabsecond}0.146 & \cellcolor{tabsecond}0.207 &  \cellcolor{tabthird}0.244 &  \cellcolor{tabthird}0.243 \\ \hline
Mini-Splatting-D                         &  \cellcolor{tabfirst}0.158 &  \cellcolor{tabfirst}0.255 &  \cellcolor{tabfirst}0.090 &  \cellcolor{tabfirst}0.169 & \cellcolor{tabsecond}0.262 &  \cellcolor{tabfirst}0.190 &  \cellcolor{tabfirst}0.172 &  \cellcolor{tabfirst}0.114 & \cellcolor{tabsecond}0.175 &  \cellcolor{tabfirst}0.100 &  \cellcolor{tabfirst}0.181 &  \cellcolor{tabfirst}0.218 &  \cellcolor{tabfirst}0.204 \\
Mini-Splatting                           &    
                  0.225 &  \cellcolor{tabthird}0.327 &                      0.150 &  \cellcolor{tabthird}0.199 &  \cellcolor{tabthird}0.312 &  \cellcolor{tabthird}0.211 &  \cellcolor{tabthird}0.199 &                      0.130 &  
                    0.199 &
   0.159 &                      0.245 &       
               0.257 &                      0.250

    \end{tabular}
    }
\end{table*}

\clearpage

%
%
\bibliographystyle{splncs04}
\bibliography{egbib}

\begin{thebibliography}{10}
\providecommand{\url}[1]{\texttt{#1}}
\providecommand{\urlprefix}{URL }
\providecommand{\doi}[1]{https://doi.org/#1}

\bibitem{barron2021mipnerf}
Barron, J.T., Mildenhall, B., Tancik, M., Hedman, P., Martin-Brualla, R., Srinivasan, P.P.: {Mip-NeRF: A Multiscale Representation for Anti-Aliasing Neural Radiance Fields}. ICCV  (2021)

\bibitem{barron2022mipnerf360}
Barron, J.T., Mildenhall, B., Verbin, D., Srinivasan, P.P., Hedman, P.: {Mip-{NeRF} 360: Unbounded Anti-Aliased Neural Radiance Fields}. CVPR  (2022)

\bibitem{barron2023zip}
Barron, J.T., Mildenhall, B., Verbin, D., Srinivasan, P.P., Hedman, P.: {Zip-NeRF: Anti-Aliased Grid-Based Neural Radiance Fields}. ICCV  (2023)

\bibitem{chen2024survey}
Chen, G., Wang, W.: {A Survey on 3D Gaussian Splatting}. arXiv preprint arXiv:2401.03890  (2024)

\bibitem{chen2017fast}
Chen, S., Tian, D., Feng, C., Vetro, A., Kova{\v{c}}evi{\'c}, J.: {Fast Resampling of 3D Point Clouds via Graphs}. IEEE Transactions on Signal Processing  \textbf{66}(3),  666--681 (2017)

\bibitem{de2016compression}
De~Queiroz, R.L., Chou, P.A.: {Compression of 3D Point Clouds Using a Region-Adaptive Hierarchical Transform}. IEEE TIP  \textbf{25}(8),  3947--3956 (2016)

\bibitem{deng2023compressing}
Deng, C.L., Tartaglione, E.: {Compressing Explicit Voxel Grid Representations: fast NeRFs become also small}. In: WACV. pp. 1236--1245 (2023)

\bibitem{deng2022depth}
Deng, K., Liu, A., Zhu, J.Y., Ramanan, D.: {Depth-supervised NeRF: Fewer Views and Faster Training for Free}. In: CVPR. pp. 12882--12891 (2022)

\bibitem{dovrat2019learning}
Dovrat, O., Lang, I., Avidan, S.: Learning to sample. In: CVPR. pp. 2760--2769 (2019)

\bibitem{durvasula2023distwar}
Durvasula, S., Zhao, A., Chen, F., Liang, R., Sanjaya, P.K., Vijaykumar, N.: {DISTWAR: Fast Differentiable Rendering on Raster-based Rendering Pipelines}. arXiv preprint arXiv:2401.05345  (2023)

\bibitem{fan2023lightgaussian}
Fan, Z., Wang, K., Wen, K., Zhu, Z., Xu, D., Wang, Z.: {LightGaussian: Unbounded 3D Gaussian Compression with 15x Reduction and 200+ FPS}. arXiv preprint arXiv:2311.17245  (2023)

\bibitem{franke2024trips}
Franke, L., R{\"u}ckert, D., Fink, L., Stamminger, M.: {TRIPS: Trilinear Point Splatting for Real-Time Radiance Field Rendering}. arXiv preprint arXiv:2401.06003  (2024)

\bibitem{graham2017submanifold}
Graham, B., Van~der Maaten, L.: {Submanifold Sparse Convolutional Networks}. arXiv preprint arXiv:1706.01307  (2017)

\bibitem{hedman2018deep}
Hedman, P., Philip, J., Price, T., Frahm, J.M., Drettakis, G., Brostow, G.: {Deep Blending for Free-Viewpoint Image-Based Rendering}. SIGGRAPH Asia  (2018)

\bibitem{hu2020randla}
Hu, Q., Yang, B., Xie, L., Rosa, S., Guo, Y., Wang, Z., Trigoni, N., Markham, A.: {RandLA-Net: Efficient Semantic Segmentation of Large-Scale Point Clouds}. In: CVPR. pp. 11108--11117 (2020)

\bibitem{katz2013improving}
Katz, S., Tal, A.: {Improving the Visual Comprehension of Point Sets}. In: CVPR. pp. 121--128 (2013)

\bibitem{kerbl20233d}
Kerbl, B., Kopanas, G., Leimk{\"u}hler, T., Drettakis, G.: {3D Gaussian Splatting for Real-Time Radiance Field Rendering}. TOG  \textbf{42}(4) (2023)

\bibitem{Knapitsch2017}
Knapitsch, A., Park, J., Zhou, Q.Y., Koltun, V.: {Tanks and Temples: Benchmarking Large-Scale Scene Reconstruction}. TOG  \textbf{36}(4) (2017)

\bibitem{lang2020samplenet}
Lang, I., Manor, A., Avidan, S.: {SampleNet: Differentiable Point Cloud Sampling}. In: CVPR. pp. 7578--7588 (2020)

\bibitem{lee2023compact}
Lee, J.C., Rho, D., Sun, X., Ko, J.H., Park, E.: {Compact 3D Gaussian Representation for Radiance Field}. arXiv preprint arXiv:2311.13681  (2023)

\bibitem{li2023compressing}
Li, L., Shen, Z., Wang, Z., Shen, L., Bo, L.: {Compressing Volumetric Radiance Fields to 1 MB}. In: CVPR. pp. 4222--4231 (2023)

\bibitem{luiten2023dynamic}
Luiten, J., Kopanas, G., Leibe, B., Ramanan, D.: {Dynamic 3D Gaussians: Tracking by Persistent Dynamic View Synthesis}. 3DV  (2024)

\bibitem{lv2022intrinsic}
Lv, C., Lin, W., Zhao, B.: {Intrinsic and Isotropic Resampling for 3D Point Clouds}. PAMI  \textbf{45}(3),  3274--3291 (2022)

\bibitem{mildenhall2020}
Mildenhall, B., Srinivasan, P.P., Tancik, M., Barron, J.T., Ramamoorthi, R., Ng, R.: {NeRF: Representing Scenes as Neural Radiance Fields for View Synthesis}. ECCV  (2020)

\bibitem{moenning2004intrinsic}
Moenning, C., Dodgson, N.A.: {Intrinsic Point Cloud Simplification}. Proc. 14th GrahiCon  \textbf{14}(23), ~2 (2004)

\bibitem{muller2022instant}
M\"uller, T., Evans, A., Schied, C., Keller, A.: {Instant Neural Graphics Primitives with a Multiresolution Hash Encoding}. SIGGRAPH  (2022)

\bibitem{niedermayr2023compressed}
Niedermayr, S., Stumpfegger, J., Westermann, R.: {Compressed 3D Gaussian Splatting for Accelerated Novel View Synthesis}. arXiv preprint arXiv:2401.02436  (2023)

\bibitem{niemeyer2024radsplat}
Niemeyer, M., Manhardt, F., Rakotosaona, M.J., Oechsle, M., Duckworth, D., Gosula, R., Tateno, K., Bates, J., Kaeser, D., Tombari, F.: {RadSplat: Radiance Field-Informed Gaussian Splatting for Robust Real-Time Rendering with 900+ FPS}. arXiv preprint arXiv:2403.13806  (2024)

\bibitem{pauly2002efficient}
Pauly, M., Gross, M., Kobbelt, L.P.: {Efficient Simplification of Point-Sampled Surfaces}. In: VIS. pp. 163--170. IEEE (2002)

\bibitem{qi2017pointnet}
Qi, C.R., Su, H., Mo, K., Guibas, L.J.: {PointNet: Deep Learning on Point Sets for 3D Classification and Segmentation}. In: CVPR. pp. 652--660 (2017)

\bibitem{qian2023gaussianavatars}
Qian, S., Kirschstein, T., Schoneveld, L., Davoli, D., Giebenhain, S., Nie{\ss}ner, M.: {GaussianAvatars: Photorealistic Head Avatars with Rigged 3D Gaussians}. CVPR  (2024)

\bibitem{rho2023masked}
Rho, D., Lee, B., Nam, S., Lee, J.C., Ko, J.H., Park, E.: {Masked Wavelet Representation for Compact Neural Radiance Fields}. In: CVPR. pp. 20680--20690 (2023)

\bibitem{schoenberger2016mvs}
Sch\"{o}nberger, J.L., Zheng, E., Pollefeys, M., Frahm, J.M.: {Pixelwise View Selection for Unstructured Multi-View Stereo}. In: ECCV (2016)

\bibitem{sun2022direct}
Sun, C., Sun, M., Chen, H.T.: {Direct Voxel Grid Optimization: Super-fast Convergence for Radiance Fields Reconstruction}. CVPR  (2022)

\bibitem{sun2024recent}
Sun, J.M., Wu, T., Gao, L.: {Recent Advances in Implicit Representation-based 3D Shape Generation}. Visual Intelligence  \textbf{2}(1), ~9 (2024)

\bibitem{tang2023dreamgaussian}
Tang, J., Ren, J., Zhou, H., Liu, Z., Zeng, G.: {DreamGaussian: Generative Gaussian Splatting for Efficient 3D Content Creation}. ICLR  (2024)

\bibitem{wen2023learnable}
Wen, C., Yu, B., Tao, D.: {Learnable Skeleton-Aware 3D Point Cloud Sampling}. In: CVPR. pp. 17671--17681 (2023)

\bibitem{xie2023hollownerf}
Xie, X., Gherardi, R., Pan, Z., Huang, S.: {HollowNeRF: Pruning Hashgrid-Based NeRFs with Trainable Collision Mitigation}. In: ICCV. pp. 3480--3490 (2023)

\bibitem{xu2022point}
Xu, Q., Xu, Z., Philip, J., Bi, S., Shu, Z., Sunkavalli, K., Neumann, U.: {Point-NeRF: Point-based Neural Radiance Fields}. In: CVPR. pp. 5438--5448 (2022)

\bibitem{yan2024street}
Yan, Y., Lin, H., Zhou, C., Wang, W., Sun, H., Zhan, K., Lang, X., Zhou, X., Peng, S.: {Street Gaussians for Modeling Dynamic Urban Scenes}. arXiv preprint arXiv:2401.01339  (2024)

\bibitem{yu2021plenoxels}
Yu, A., Fridovich-Keil, S., Tancik, M., Chen, Q., Recht, B., Kanazawa, A.: {Plenoxels: Radiance Fields without Neural Networks}. CVPR  (2022)

\bibitem{yu2023mip}
Yu, Z., Chen, A., Huang, B., Sattler, T., Geiger, A.: {Mip-Splatting: Alias-free 3D Gaussian Splatting}. CVPR  (2024)

\bibitem{zwicker2002ewa}
Zwicker, M., Pfister, H., Van~Baar, J., Gross, M.: {EWA Splatting}. TVCG  \textbf{8}(3),  223--238 (2002)

\end{thebibliography}

\end{document}